\documentclass{article}

\usepackage{arxiv}

\usepackage[utf8]{inputenc} 
\usepackage[T1]{fontenc}    
\usepackage{hyperref}       
\usepackage{url}            
\usepackage{booktabs}       
\usepackage{amsfonts}       
\usepackage{nicefrac}       
\usepackage{microtype}      
\usepackage{lipsum}		
\usepackage{graphicx}
\usepackage[numbers]{natbib}
\usepackage{doi}
\usepackage[linesnumbered,ruled,vlined]{algorithm2e}
\usepackage{subcaption}
\usepackage{amsmath}

\usepackage[labelfont=bf,labelsep=period]{caption}

\title{Enhancing Leaf Disease Classification Using GAT-GCN Hybrid Model}

\author{ \href{https://orcid.org/0009-0003-8930-1204}{\includegraphics[scale=0.06]{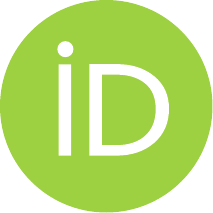}\hspace{1mm}Shyam Sundhar}\\
	School of Computer Science and Engineering \\
	Vellore Institute of Technology\\
	Chennai, Tamilnadu, India  \\
	\texttt{} \\
	\And
	{Riya Sharma} \\
	School of Computer Science and Engineering\\
	Vellore Institute of Technology\\
	Chennai, Tamilnadu, India \\
	\texttt{} \\
    \And
    {Priyansh Maheshwari} \\
	School of Computer Science and Engineering\\
	Vellore Institute of Technology\\
	Chennai, Tamilnadu, India \\
	\texttt{} \\
    \And
    \href{https://orcid.org/0009-0009-0141-2617}
    {\includegraphics[scale=0.06]{orcid.pdf}\hspace{1mm}Suvidha Rupesh Kumar}\thanks{Corresponding author. Email: suvidha.rupesh@vit.ac.in} \\
	School of Computer Science and Engineering\\
	Vellore Institute of Technology\\
	Chennai, Tamilnadu, India \\
	\texttt{} \\
    \And
    \href{https://orcid.org/0000-0003-0934-7230}
    {\includegraphics[scale=0.06]{orcid.pdf}\hspace{1mm}T. Sunil Kumar} \\
	Department of Electrical Engineering, Mathematics and Science\\
	University of Gavle, \\
	Gavle, Sweden \\
	\texttt{} \\
}

\date{}

\renewcommand{\shorttitle}{Enhancing Leaf Disease Classification Using GAT-GCN Hybrid Model}

\hypersetup{
pdftitle={A template for the arxiv style},
pdfsubject={q-bio.NC, q-bio.QM},
pdfauthor={David S.~Hippocampus, Elias D.~Striatum},
pdfkeywords={First keyword, Second keyword, More},
}

\begin{document}
\maketitle

\begin{abstract}
Agriculture plays a critical role in the global economy, providing livelihoods and ensuring food security for billions. As innovative agricultural practices become more widespread, the risk of crop diseases has increased, highlighting the urgent need for efficient, low-intervention disease identification methods. This research presents a hybrid model combining Graph Attention Networks (GATs) and Graph Convolution Networks (GCNs) for leaf disease classification. GCNs have been widely used for learning from graph-structured data, and GATs enhance this by incorporating attention mechanisms to focus on the most important neighbors. The methodology integrates superpixel segmentation for efficient feature extraction, partitioning images into meaningful, homogeneous regions that better capture localized features. The authors have employed an edge augmentation technique to enhance the robustness of the model. The edge augmentation technique has introduced a significant degree of generalization in the detection capabilities of the model. To further optimize training, weight initialization techniques are applied. The hybrid model is evaluated against the individual performance of the GCN and GAT models and the hybrid model achieved a precision of 0.9822, recall of 0.9818, and F1-score of 0.9818 in apple leaf disease classification, a precision of 0.9746, recall of 0.9744, and F1-score of 0.9743 in potato leaf disease classification, and a precision of 0.8801, recall of 0.8801, and F1-score of 0.8799 in sugarcane leaf disease classification. These results demonstrate the robustness and performance of the model, suggesting its potential to support sustainable agricultural practices through precise and effective disease detection. This work is a small step towards reducing the loss of crops and hence supporting sustainable goals of zero hunger and life on land. 
\end{abstract}


\section{Introduction}
The detection of plant diseases is important in agriculture, significantly impacting crop yield and overall productivity. Plant disease leads to biological and economic losses that leave millions of people starving and undernourished \cite{oerke2006crop}. As agricultural practices grow more complex and climate change intensifies, the prevalence of crop diseases has increased, necessitating efficient and accurate identification methods. Traditional approaches often rely on manual inspections by farmers, which can be time consuming and subjective, leading to delays in disease management and substantial crop losses.

In response to the evolving demands of modern agriculture, there is an urgent need for effective disease detection strategies. Plant disease, especially due to infected leaves, can be identified by the change in leaf color or other changes such as leaf shrinkage \cite{agrios2005plant}. This visual clue has paved the way for identification using computer vision methods. The integration of machine learning and advanced computer vision technologies offers a promising solution \cite{jafar2024revolutionizing}, enabling rapid analysis of leaf images and facilitating timely interventions essential for maintaining healthy crops.

Recent advances in machine learning and deep learning have significantly improved the efficiency of plant disease detection. For example, studies have shown the effectiveness of computer vision algorithms for the detection of tomato leaf disease \cite{harakannanavar2022plant}, while the application of deep learning architectures such as AlexNet has shown potential for the detection of olive disease \cite{alruwaili2019efficient}. 

However, despite the capabilities of modern deep learning (DL) and machine learning (ML) techniques, challenges remain to capture the diverse characteristics inherent in crops due to variations in environmental conditions, lighting, and plant physiology. This limitation highlights the need for a generalized model that can effectively adapt to these variations and accurately identify diseases across different contexts.

To address these challenges, this research article proposes a hybrid model that integrates Graph Attention Networks (GAT) and Graph Convolution Networks (GCN) for leaf disease classification. GCN and GAT-based architectures have been effectively applied in various domains, including drug discovery \cite{sun2020graph}, where GCNs model molecular interactions, 3D shape analysis \cite{wei2020view}, where a view-based GCN aggregates multi-view features for shape recognition, and point cloud segmentation \cite{wang2019graph}, where a Graph Attention Convolution (GAC) dynamically adjusts its receptive field to capture fine-grained object structures. Inspired by these applications, this study applies GCN-GAT to leaf disease classification, leveraging ability of GCN to model spatial correlations among leaf features and attention mechanism of GCN to prioritize critical patterns indicative of disease symptoms. This integration aims to enhance classification accuracy and robustness, particularly in varying environmental conditions.

In summary, the major contributions of this research includes:
\begin{itemize}
    \item The GCN-GAT hybrid model, integrating spatial feature-capturing of GCN and feature prioritization strengths of GAT.
    \item Utilized image segmentation, edge augmentation, and weight initialization to enhance architectural robustness.
    \item Analyzed performance on three diverse datasets (apple, potato, and sugarcane leaves), showcasing model adaptability.
    \item Conducted comparative experiments with standalone GCN and GAT models to validate the effectiveness of the architecture.
    \item Evaluated using key metrics like F1-score, accuracy, precision, recall, cross-entropy loss, and confusion matrix.
\end{itemize}

\section{Review of prior findings}

This literature review aims to provide an overview of the recent advancements in the realm of plant leaf disease detection. It examines a range of approaches that involves traditional machine learning approaches, state-of-the-art deep learning architectures that integrate multiple techniques. By synthesizing findings from key studies, the review also identifies research gaps and highlights areas that require further exploration, particularly in enhancing model adaptability and generalizability as well as the improvement of disease detection systems in diverse and dynamic agricultural environments.

Oo et al. (2018) \cite{oo2018plant} presented a detection and classification system for four plant leaf diseases, including Cercospora Leaf Spot, Bacterial Blight, Powdery Mildew, and Rust. It involves image preprocessing, segmentation, and feature extraction using GLCM and LBP techniques. Various machine learning models were used, with the highest accuracy of 98.2\% achieved by SVM. A comparative evaluation was conducted using classifiers such as SVM, KNN, and ensemble methods. The use of texture feature extraction through GLCM and LBP enhanced classification by leveraging both statistical and structural characteristics. However, the study relies on feature extraction techniques that lack structural relationships among the features of the leaves.

Karlekar et al. (2020) \cite{karlekar2020soynet} addressed issues related to leaf segmentation and disease classification with a two-part approach: extracting leaf images through a specific method and employing a deep learning model called SoyNet for soybean disease classification. The model achieved an accuracy of 98.14\%. An Integrated Pest Management (IPM) technique ensured the segmentation of leaf regions, even in complex backgrounds, focusing the model on relevant areas. Additional testing on the large Plant Disease Database (PDDB) dataset, which contains 16 disease categories, aimed to improve generalization across various disease types. However, since the model was tested on only one dataset, its adaptability to new data remains limited.

Wang et al. (2022) \cite{wang2022plant} proposed a plant disease recognition model integrating visual and textual data through feature decomposition and GCNs. The model was evaluated on datasets with both uniform and non-uniform severity levels. Traditional networks like ResNet18 performed well on uniform data, while feature decomposition improved results for more complex data. The reported accuracy was 97.62\%, with precision, sensitivity, and specificity values of 92.81\%, 98.54\%, and 93.57\%, respectively, demonstrating the efficacy of multimodal approaches. It was observed that two-layer GCNs outperformed one-layer and three-layer models by optimally extracting features from the graph structure. However, the study utilized a static graph structure, which may not fully capture the dynamic nature of disease progression. Dynamic graph models could be explored to better represent changing disease information.

A review by Lu et al. (2021) \cite{lu2021review} focused on the application of deep learning methods, specifically CNNs, for plant leaf disease classification. Techniques like segmentation, data augmentation, and transfer learning were examined to overcome challenges such as limited datasets and robustness issues. Transfer learning-based CNNs achieved up to 95\% accuracy. The review also explored image segmentation techniques like watershed segmentation, Otsu's thresholding, and K-means clustering to isolate leaves from complex backgrounds—an often overlooked aspect in related research. While CNNs excel at capturing pixel-level features, they lack the capability to model relational and structural information inherent in disease patterns.

Rao et al. (2024) \cite{rao2024plant} presented a study combining CNNs and GCNs for improved plant disease classification. The integrated model achieved 99\% accuracy, surpassing traditional models like DeepPlantNet (98\%), ensemble models (91\%), and transfer learning approaches (95\%). By incorporating image features and plant connectivity, the model provides better contextual awareness and higher accuracy in disease classification. However, this approach requires significant computational resources and longer training times compared to standalone CNN or GCN models. Additionally, its performance heavily depends on the availability of large and balanced datasets.

Peng et al. (2022) \cite{peng2022leaf} introduced a new system for the automatic identification, localization, and detection of leaf diseases using an image retrieval approach that incorporates object detection and deep metric learning. They first enhanced the YOLOv5 algorithm to improve the detection of small objects, allowing for more accurate extraction of leaf objects. The system also integrates classification recognition with metric learning, enabling the model to jointly learn both categorization and similarity measurements, thereby enhancing the performance of existing image classification models. This approach allows for the addition of new disease types without the need for retraining. Experimental results on three publicly available leaf disease datasets demonstrated the effectiveness of the proposed system. The study highlights the system's potential for practical applications in intelligent agriculture, such as crop health monitoring and nutrition diagnosis. However, there is a need to improve the  scalability and adaptability of the system to handle a broader range of plant species and environmental variations, ensuring its robustness in diverse agricultural settings.

Rathore et al. (2020) \cite{rathore2020automatic} proposed an automatic method for detecting rice plant diseases using a CNN. The model classified rice images into "Healthy" and "Leaf Blast" categories with an accuracy of 99.61\% using a dataset of 1,000 images. Data augmentation techniques, including random rotation, shifting, flipping, and cropping, were employed to expand the dataset and improve the generalization of the model. While effective, the applicability of the model is limited, as it only classifies two classes—healthy and leaf blast—restricting its use in detecting a broader range of rice diseases.

Roy et al. (2023) \cite{roy2023disease} introduced an advanced plant disease segmentation method for precision agriculture using optimal dimensionality reduction with fuzzy C-means clustering and deep learning. The approach focused on segmenting rice leaf regions and classifying diseases with high accuracy. The model utilized CNNs and achieved an accuracy of 99.61\% on a dataset containing 1,000 images. Data augmentation techniques were employed to enhance performance. However, similar to Rathore et al. (2020) \cite{rathore2020automatic}, the model was designed to classify only two categories: healthy and leaf blast, thereby limiting its generalizability to a wider variety of diseases and pests.

Liu et al. (2020) \cite{liu2020grape} proposed a novel approach for identifying grape leaf diseases using an improved convolution neural network (CNN). The study focused on six major grape leaf diseases—anthracnose, brown spot, mites, black rot, downy mildew, and leaf blight—that cause significant economic losses in the grape industry. To address this challenge, the authors developed a dataset of 107,366 grape leaf images through image enhancement techniques, utilizing 4,023 field-collected images and 3,646 images from public datasets. The method incorporates an Inception structure to enhance multi-dimensional feature extraction and introduces a dense connectivity strategy to promote feature reuse and propagation. The proposed deep learning model, named DICNN, was trained from scratch and achieved an overall accuracy of 97.22\% on a hold-out test set. Compared to GoogLeNet and ResNet-34, DICNN improved recognition accuracy by 2.97\% and 2.55\%, respectively. While the model demonstrates strong performance, a key research gap remains in its ability to generalize across different crops and diverse environmental conditions

Bansal et al. (2021) \cite{bansal2021disease} developed a deep learning-based approach for detecting apple leaf diseases. The dataset consisted of 3,642 images divided into four classes: apple scab, apple cedar rust, multiple diseases, and healthy leaves. Preprocessing techniques such as flipping, rotation, and blurring were applied, and images were resized to 512x512 pixels. The authors employed pre-trained models, including DenseNet121, EfficientNetB7, and EfficientNet NoisyStudent, with an ensemble model achieving a maximum accuracy of 96.25\%. The ensemble approach effectively reduced variance and improved classification accuracy. However, relying solely on pre-trained CNN models may not fully capture the intricate relationships between different parts of a leaf, especially in cases involving multiple diseases.

Dai et al. (2022) \cite{dai2022graphcda} proposed a hybrid graph representation learning framework, GraphCDA, which combines GCN and GAT for predicting disease-associated circRNAs. The study highlights the importance of automated model selection in production systems through online experimentation. A Bayesian surrogate model was employed to derive the probability distribution of the metric of interest, guiding efficient model selection. This approach effectively balances exploration and exploitation, resulting in improved performance. Although the framework focuses on the prediction of circRNA, its graph-based learning techniques and automated experimentation mechanisms could bring about advances in plant disease classification by allowing more dynamic and adaptive models.

Liu et al. (2023) \cite{liu2023tomato} proposed a novel tomato disease object detection method combining a prior knowledge attention mechanism and multi-scale features (PKAMMF) to address challenges in tomato disease detection, such as dense target distributions, large-scale variations, and insufficient feature information for small objects in complex backgrounds. The method integrates prior knowledge with the visual features of tomato disease images via an attention mechanism, thereby enhancing feature representation. Additionally, the article introduced a new feature fusion layer in the Neck section to reduce feature loss, as well as a specialized prediction layer designed to improve the detection of small targets. A new loss function, Adaptive Structured IoU (A-SIOU), is employed to optimize bounding box regression. Experimental results on a self-built tomato disease dataset demonstrate the effectiveness of PKAMMF, achieving a mean average precision (mAP) of 91.96\%, a 3.86\% improvement over baseline methods. However, the study highlights the need for further advancements in detecting diseases under more diverse and complex environmental conditions, suggesting that future work should focus on developing more generalized models.

Khan et al. (2022) \cite{khan2022deep} proposed a two-stage apple disease detection system that achieved 88\% classification accuracy. The approach utilized transfer learning with the Xception model and Faster-RCNN for disease localization. The system was proven to be effective in detecting and localizing symptoms, including small spots, offering the advantage of enhanced detection sensitivity, even for less noticeable signs of the disease. However, a key research gap lies in improving the performance of the system in real time, as well as its ability to detect subtle symptoms in a wider range of environmental conditions, which would be crucial for practical application in agricultural settings.

Luo et al. (2021) \cite{luo2021apple} proposed an optimized multi-scale fusion network for apple leaf disease detection. Their methodology enhanced the information flow in the ResNet backbone and improved the downsampling process to retain discrete information. In addition, it incorporated pyramid convolution and dilated convolution to increase the robustness and accuracy of the model. This approach achieved a classification accuracy of 94. 24\% in the original data set and 94. 99\% in the preprocessed data set, demonstrating its potential to improve the performance of the model. However, there is a key research gap in further optimizing pyramid convolution techniques to enhance the adaptability of the model to diverse plant species and environmental conditions, which would broaden its applicability in real-world agricultural scenarios.

Wang et al. (2024) \cite{wang2024tomato} introduced a tomato leaf disease detection method that leverages attention mechanisms and multi-scale feature fusion to address challenges posed by varying environmental conditions in image captures. By incorporating the Convolution Block Attention Module (CBAM) into the backbone feature extraction network, the method improves lesion feature extraction and mitigates the impact of environmental interference. Additionally, the integration of shallow feature maps into a re-parameterized generalized feature pyramid network (RepGFPN) led to the development of the BiRepGFPN, which enhances feature fusion and localization of small lesions. This innovation further replaces the Path Aggregation Feature Pyramid Network (PAFPN) in YOLOv6, enabling more effective integration of deep semantic and shallow spatial information. Experimental results on the PlantDoc and tomato leaf disease datasets demonstrate significant improvements in mean average precision (mAP), precision, recall, and F1 score, showcasing the model's robust detection performance and strong potential for generalization. However, the study also indicates that while the model performs well on controlled datasets, it faces challenges when deployed in real-world settings where environmental variations and image quality fluctuations are more pronounced. 

Mahum et al. (2023) \cite{mahum2023novel} proposed an improved deep-learning technique for the classification of potato leaf diseases. Their approach utilized a pre-trained Efficient DenseNet model with an additional transition layer in DenseNet-201 and incorporated a reweighted cross-entropy loss function to address the issue of imbalanced data. The proposed algorithm achieved an accuracy of 97.2\%, outperforming existing methods and demonstrating the effectiveness of leveraging pre-trained models in handling imbalanced datasets. However, the major research gap lies in enhancing the ability of the model to detect rare disease instances, particularly when data for these conditions is limited, which would improve its robustness in real-world applications.

Bera et al. (2024) \cite{bera2024pnd} proposed a deep learning method  called Plant Nutrition Deficiency and Disease Network (PND-Net) for classifying plant nutrition deficiencies and diseases. Their approach utilized a graph convolution network (GCN) added on top of a convolution neural network (CNN), and introduced region-based feature summarization through spatial pyramidal pooling to enhance feature representation. The model achieved classification accuracies of 90.00\% for banana nutrition deficiency, 90.54\% for coffee nutrition deficiency, 96.18\% for potato diseases, and 84.30\% for PlantDoc diseases, demonstrating its strong performance across various plant categories. Despite these promising results, a research gap remains in improving the scalability of the model for multi-label classification tasks, particularly when dealing with more complex and diverse plant disease datasets, which would be essential for broader agricultural applications.

Shoaib et al. (2022) \cite{shoaib2022deep} proposed a deep learning-based system for detecting tomato plant diseases using images of plant leaves. Their model utilizes an Inception Net architecture, trained on a dataset of 18,161 segmented and non-segmented images. To detect and segment disease-affected regions, the authors employed two state-of-the-art semantic segmentation models—U-Net and Modified U-Net. The study demonstrates that the Modified U-Net model outperforms the basic U-Net model with an accuracy of 98.66\%, an IoU score of 98.5, and a Dice score of 98.73\%. InceptionNet achieved remarkable accuracy, with 99.95\% for binary classification and 99.12\% for six-class classification, surpassing the performance of the Modified U-Net model. While the proposed method outperforms existing techniques in tomato disease classification, a significant research gap lies in enhancing the ability of the model to handle the detection and segmentation of plant diseases under real-world conditions, such as varying lighting, leaf orientations, and occlusions.

In contrast to existing architectures and methodologies which integrates a hybrid GCN-GAT model, effectively tackles the challenges of generalization and spatial relationship-based feature extraction. This ensures enhanced and robust performance in the domain of leaf disease detection. The key research gap in this field can be summarized as follows:
\begin{itemize} 
\item Existing state-of-the-art architectures struggle to generalize across diverse plant species and disease types, limiting scalability. \
\item Traditional approaches focus on pixel-level features and fail to capture spatial and relational dependencies between leaf regions, reducing accuracy. 
\item Many state-of-the-art architectures perform inconsistently under varying environmental conditions, such as changes in light, weather, or background noise. 
\end{itemize}

\section{Methodology}
\subsection{Dataset Description}
The hybrid model is analyzed using three different leaf disease datasets, namely: Sugarcane Leaf Disease Dataset \cite{sugarcane_leaf_disease_dataset}, Potato Leaf Disease Dataset \cite{potato_leaf_disease_dataset}, and Apple Leaf Disease Dataset \cite{apple_leaf_disease_dataset}, each capturing a diverse set of images taken under various lighting conditions.

\subsubsection{Sugarcane Leaf Disease Dataset} The Sugarcane Leaf Disease Dataset consists of 2521 RGB images of sugarcane leaves, manually collected from various regions in Maharashtra, India. The images are categorized into five distinct classes, including 522 images of healthy leaves, 462 images showing mosaic symptoms, 518 images of red rot disease, 514 images of rust, and 505 images of yellow disease.

\begin{figure}[!htbp]
    \centering
    \begin{subfigure}[b]{0.45\textwidth}
        \centering
        \includegraphics[width=\linewidth]{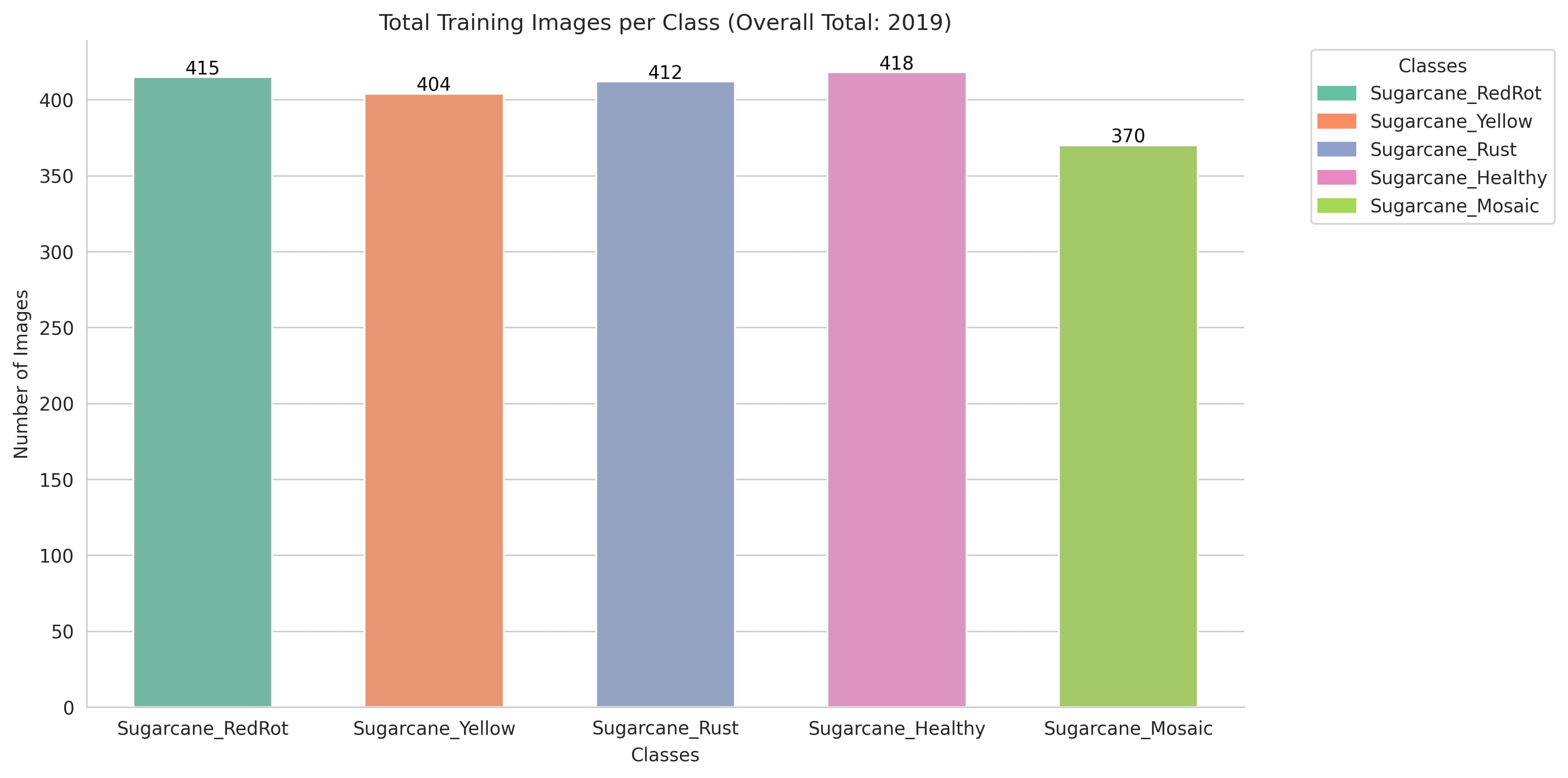}
        \caption{Training Data}
        \label{fig:sugarcane-train-data}
    \end{subfigure}
    \hfill
    \begin{subfigure}[b]{0.45\textwidth}
        \centering
        \includegraphics[width=\linewidth]{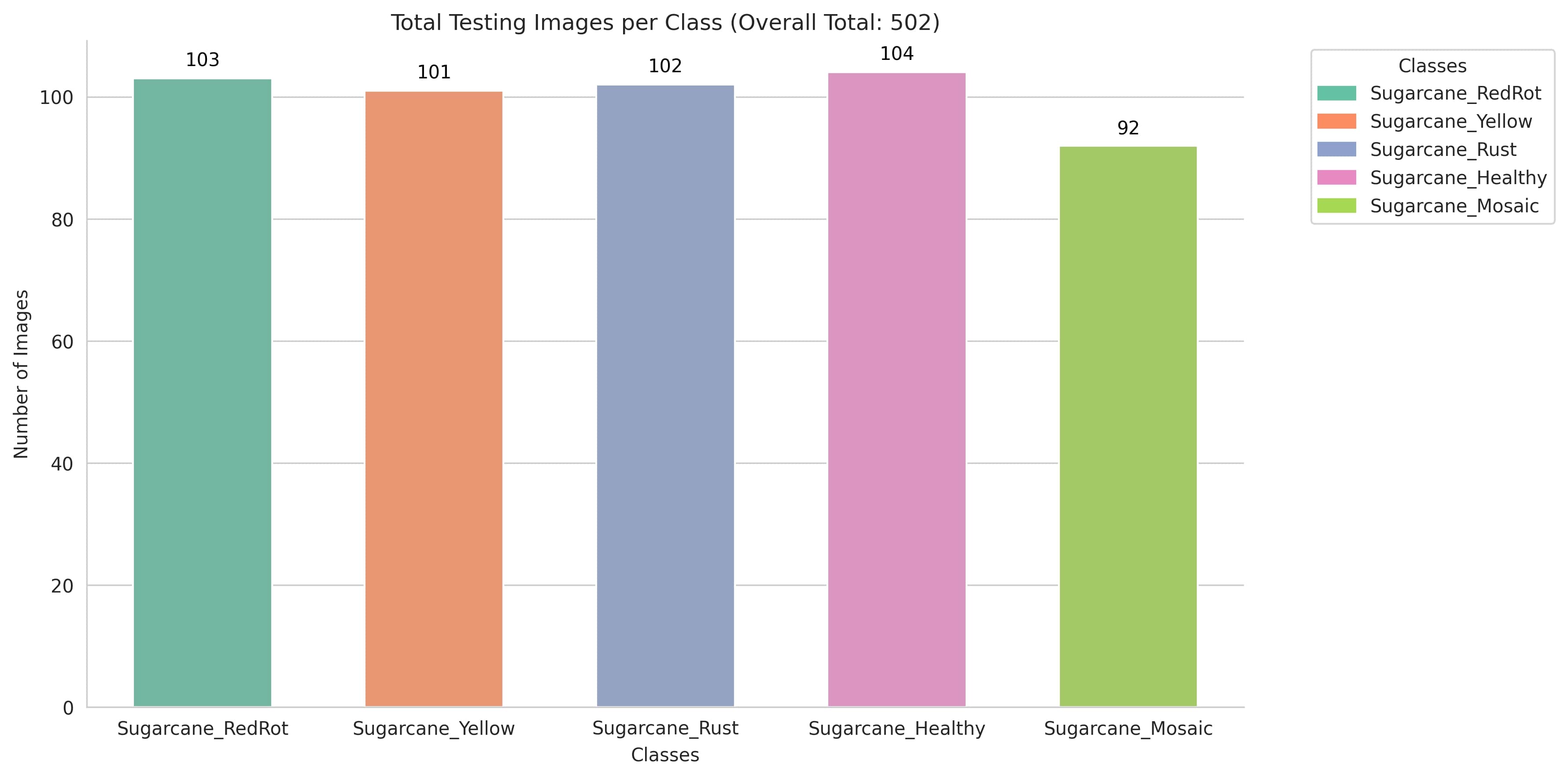}
        \caption{Test Data}
        \label{fig:sugarcane-test-data}
    \end{subfigure}
    \caption{Sugarcane Leaf Dataset distribution among train-test splits: (a) Distribution of Training Data, (b) Distribution of Test Data}
    \label{fig:sugarcane-dataset-distribution}
\end{figure}

\subsubsection{Potato Leaf Disease Dataset} The Potato Leaf Disease Dataset contains 1200 RGB images of potato leaves, categorized into three disease types: Early Blight, Late Blight, and Healthy leaves with each class having 300 samples of leaves.

\begin{figure}[!htbp]
    \centering
    \begin{subfigure}[b]{0.45\textwidth}
        \centering
        \includegraphics[width=\linewidth]{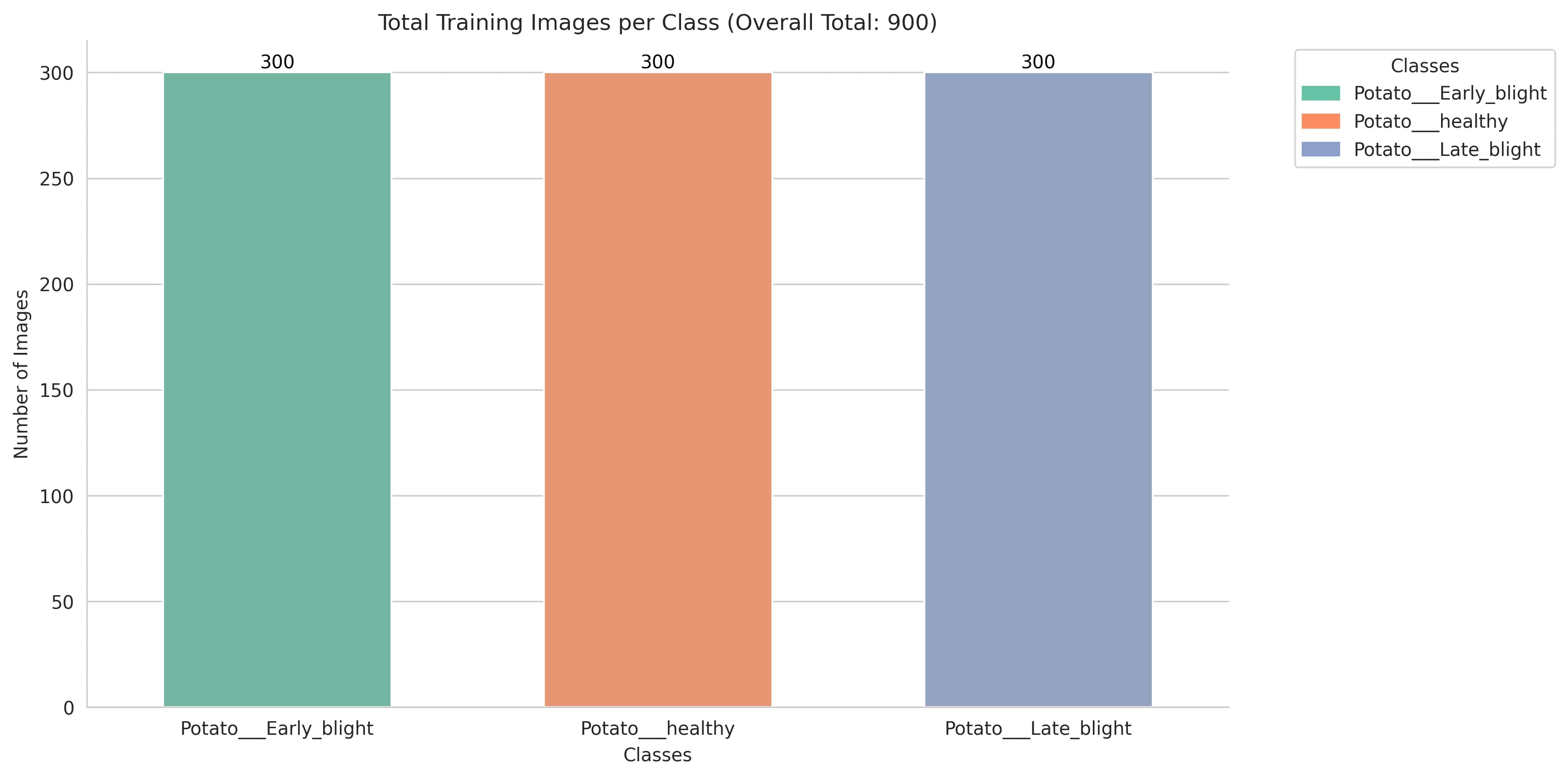}
        \caption{Training Data}
        \label{fig:potato-train-data}
    \end{subfigure}
    \hfill
    \begin{subfigure}[b]{0.45\textwidth}
        \centering
        \includegraphics[width=\linewidth]{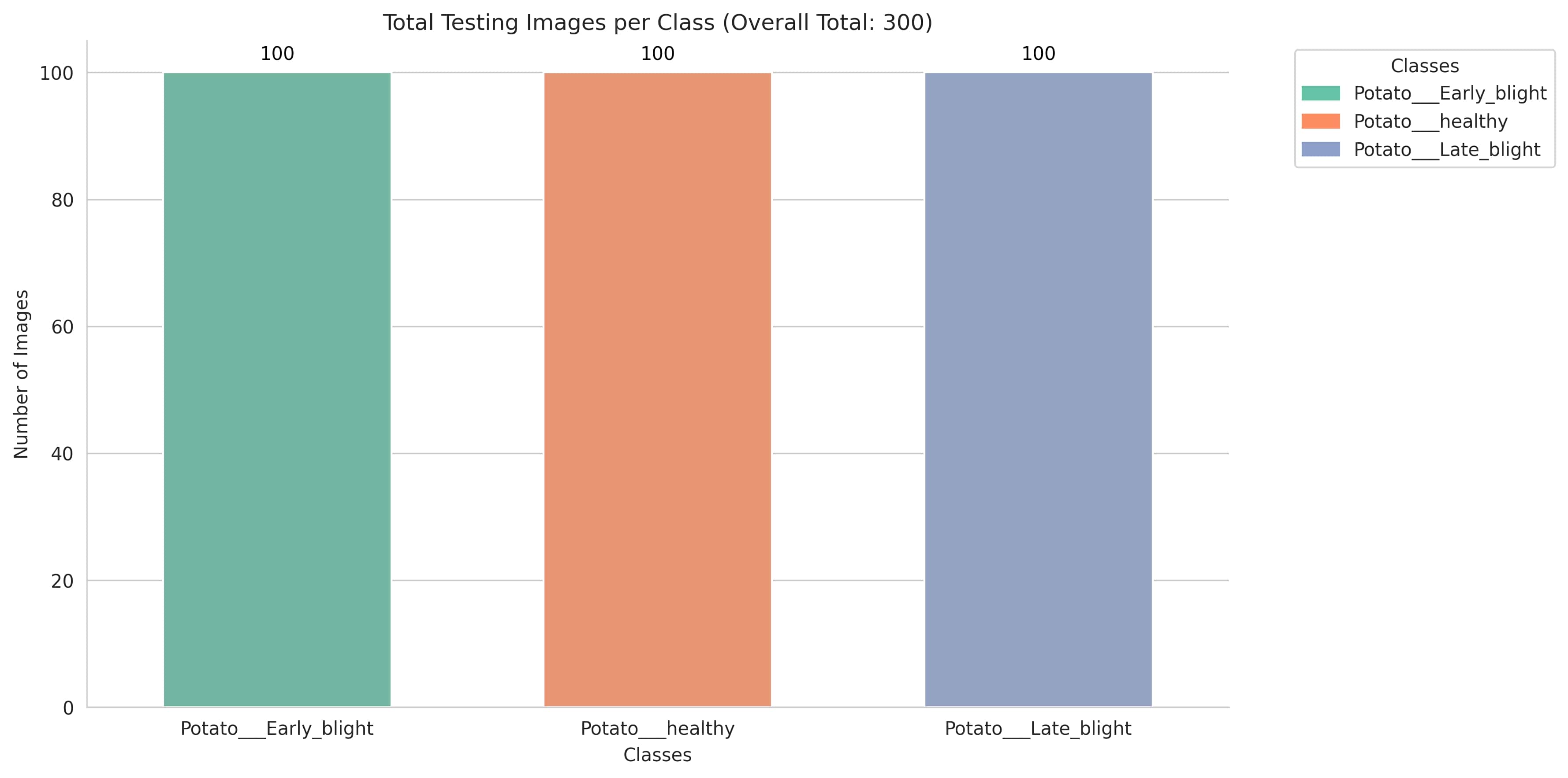}
        \caption{Test Data}
        \label{fig:potato-test-data}
    \end{subfigure}
    \caption{Potato Leaf Dataset distribution among train-test splits: (a) Distribution of Training Data, (b) Distribution of Test Data}
    \label{fig:potato-dataset-distribution}
\end{figure}

\subsubsection{Apple Leaf Diseases Dataset} The Apple Leaf Diseases Dataset is a comprehensive collection of 480 RGB images designed to support the identification of various foliar diseases in apple trees. The images in this dataset are categorized into three primary disease types: Apple Black Rot (170 samples), Cedar Rust (160 samples), and Apple Scab (150 samples).

\begin{figure}[!htbp]
    \centering
    \begin{subfigure}[b]{0.45\textwidth}
        \centering
        \includegraphics[width=\linewidth]{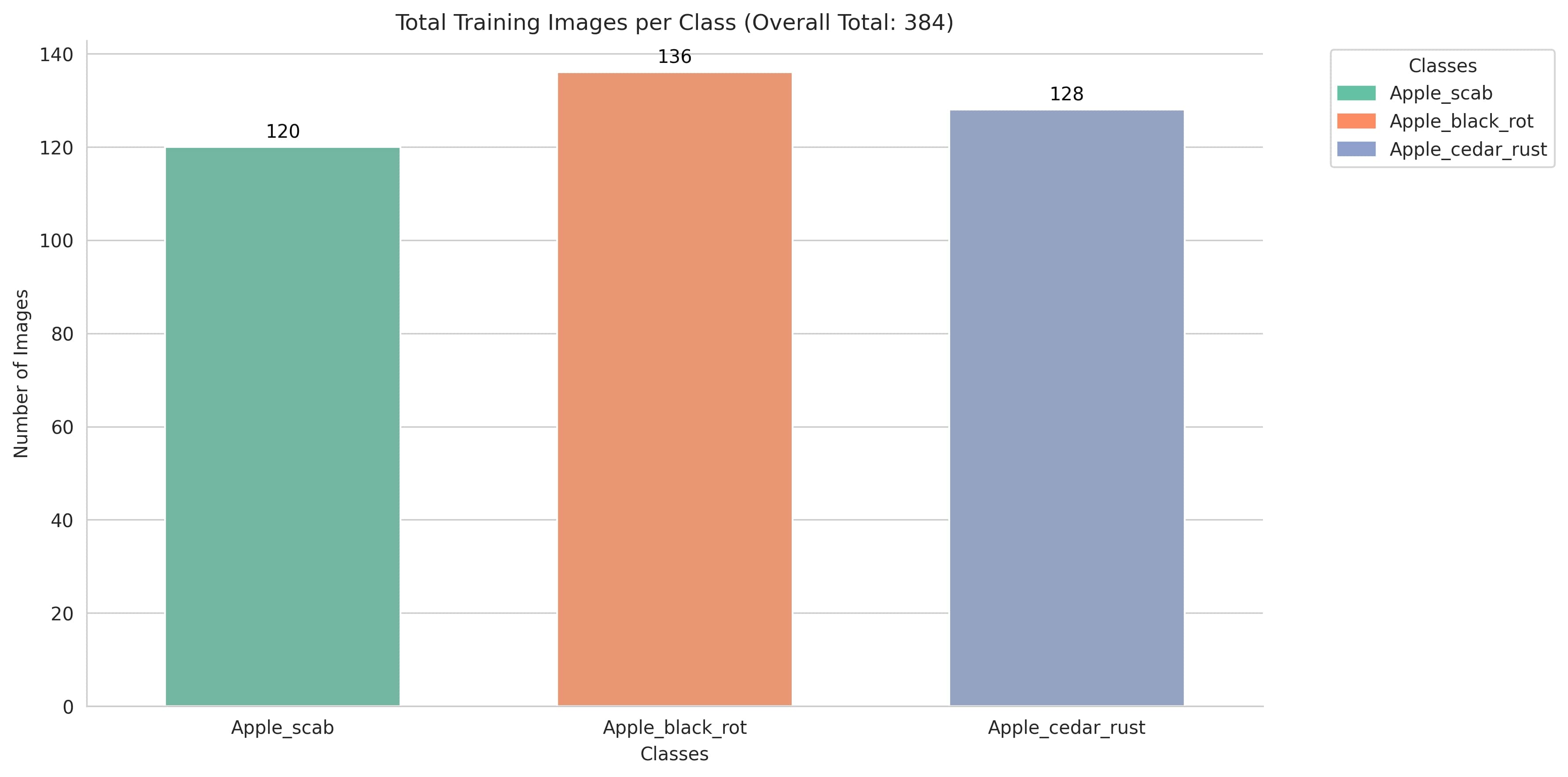}
        \caption{Training Data}
        \label{fig:apple-train-data}
    \end{subfigure}
    \hfill
    \begin{subfigure}[b]{0.45\textwidth}
        \centering
        \includegraphics[width=\linewidth]{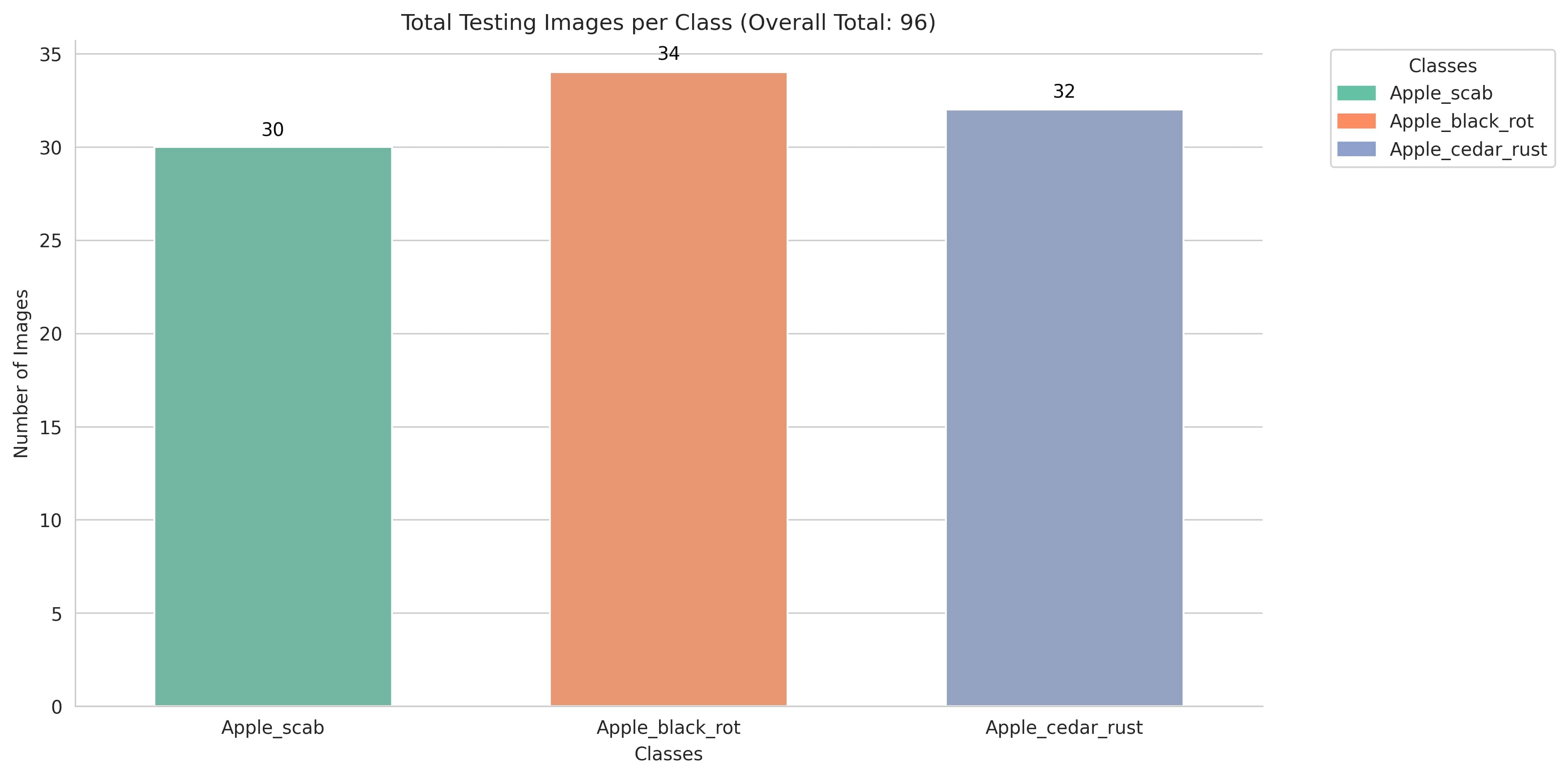}
        \caption{Test Data}
        \label{fig:apple-test-data}
    \end{subfigure}
    \caption{Apple Leaf Dataset distribution among train-test splits: (a) Distribution of Training Data, (b) Distribution of Test Data}
    \label{fig:apple-dataset-distribution}
\end{figure}

For all datasets, the images were split into training and testing sets with an 80-20 ratio to ensure a balanced representation of classes and effective model evaluation (refer to \textbf{Figs. \ref{fig:sugarcane-dataset-distribution}, \ref{fig:potato-dataset-distribution}, and \ref{fig:apple-dataset-distribution}}).

\begin{table*}[ht]
    \caption{Distribution of leaf images across classes}
    \centering
    \begin{tabular}{|p{6cm}|p{2.5cm}|p{2.5cm}|p{3cm}|}
        \hline
        \textbf{Class} & \textbf{Training} & \textbf{Testing} & \textbf{Total per class} \\
        \hline
        Sugarcane\_\_Healthy & 418 & 104 & 522 \\
        Sugarcane\_\_Mosaic & 369 & 93 & 462 \\
        Sugarcane\_\_Red Rot & 414 & 104 & 518 \\
        Sugarcane\_\_Rust & 411 & 103 & 514 \\
        Sugarcane\_\_Yellow & 404 & 101 & 505 \\
        Potato\_\_Early Blight & 300 & 100 & 400 \\
        Potato\_\_Late Blight & 300 & 100 & 400 \\
        Potato\_\_Healthy & 300 & 100 & 400 \\
        Apple\_\_Scab & 120 & 30 & 150 \\
        Apple\_\_Black Rot & 136 & 34 & 170 \\
        Apple\_\_Cedar Rust & 128 & 32 & 160 \\
        \hline
    \end{tabular}
    \label{tab:leaf-class-distribution}
\end{table*}

\subsection{Preprocessing}
To ensure the efficiency of the model, several preprocessing techniques were applied to the dataset. Initially, to standardize the input size, all images were resized to 128x128 pixels. This resizing step ensures uniformity across the dataset, which is essential for consistent model performance. Furthermore, to facilitate faster convergence during training, the pixel values of the images were normalized to a range of [-1, 1], using a mean of 0.5 and a standard deviation of 0.5. This normalization procedure helps stabilize the training process and accelerates convergence. A batch size of 32 was selected for training, balancing computational efficiency and the stability of gradient updates.

\begin{figure*}[!http]
    \centering
    \includegraphics[width=0.45\textwidth]{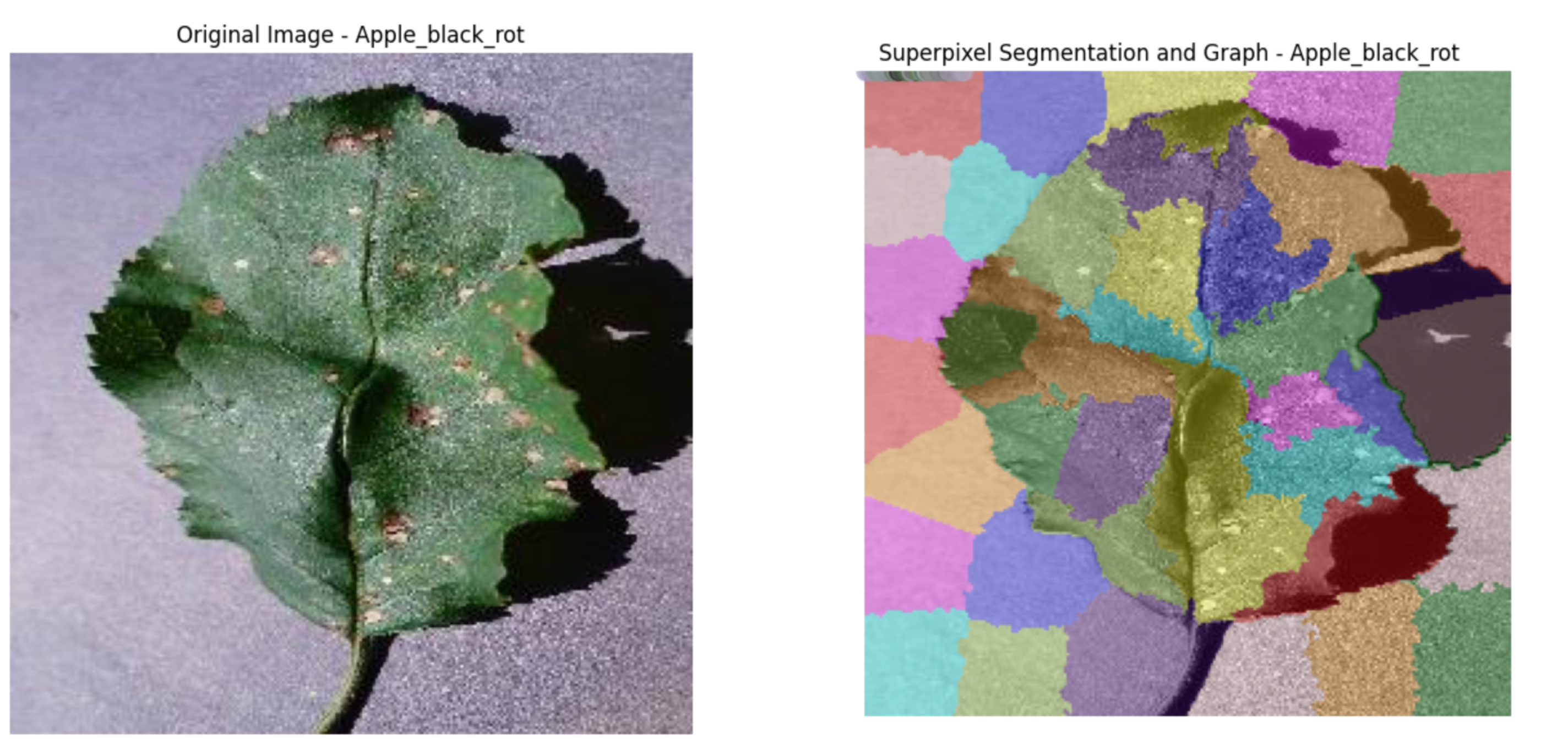}
    \caption{Preprocessed Leaf Image using Superpixel Segmentation (Apple Leaf)}
    \label{fig:sample-pre-processed-image}
\end{figure*}

In addition to resizing and normalization, superpixel segmentation was employed to enhance feature extraction. The Simple Linear Iterative Clustering (SLIC) algorithm \cite{achanta2012slic} was used to partition each image into perceptually meaningful regions, or superpixels, based on color similarity and spatial proximity. This segmentation step allows the graph models to focus on localized regions of the image, which is particularly beneficial for detecting fine-grained textures and structures that are indicative of leaf diseases. The number of segments was set to 50, providing a detailed representation of the image. This segmentation facilitates the  ability of the model to capture relevant features, improving both accuracy and generalization. The number of segments was set to 50 based on experimentation, as values of 20 segments captured fewer features, while 100 segments led to overfitting. This choice of 50 segments provided a balanced representation that enhanced feature extraction without compromising model generalization. A sample of a segmented leaf from the dataset is shown in \textbf{Fig. \ref{fig:sample-pre-processed-image}}.

Following segmentation, a Region Adjacency Graph (RAG) was constructed to capture and model the spatial relationships between adjacent superpixels. In this graph, each superpixel is represented as a node, and edges between nodes capture the adjacency relationships between superpixels. The mean RGB color of each superpixel was extracted as a node feature, offering a compact and informative representation of the visual content of the image. The adjacency information encoded in the edges enables the graph-based models to learn the spatial relationships between superpixels, enhancing its ability to recognize complex patterns in the image. These RAG features were subsequently stored in pickle files along with labels of the subsequent classes, which were utilized for training the architectures.

\subsection{Methodology}
\subsubsection{Graph Convolution Network (GCN)}
\begin{figure}[!http]
    \centering
    \includegraphics[width=0.9\textwidth]{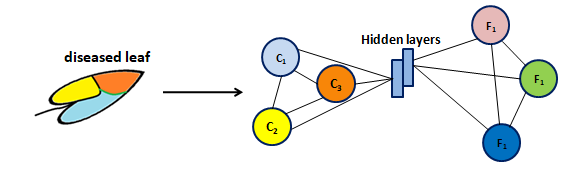}
    \caption{GCN architecture}
    \label{fig:gcn-architecture}
\end{figure}

A Graph Convolution Network (GCN) is a neural network designed for graph-structured data \cite{kipf2016semi}. It captures node-level features and local relationships between neighboring nodes by aggregating information from them. The core mechanism of GCNs involves message passing, where each node updates its representation based on the features of its neighbors. This is achieved through feature propagation, where nodes receive information from connected neighbors, followed by graph convolution, where aggregated features are transformed through a weighted sum. A sample structure of a GCN is depicted in (\textbf{Fig. \ref{fig:gcn-architecture}}).

\subsubsection{Graph Attention Network (GAT)}
\begin{figure}[!http]
    \centering
    \includegraphics[width=0.7\textwidth]{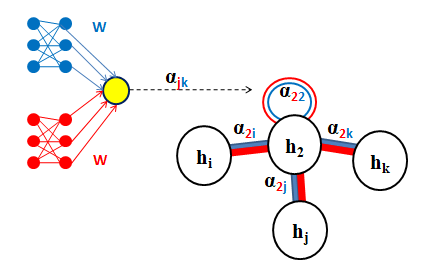}
    \caption{Graph Attention Network (GAT) architecture}
    \label{fig:gat-weight}
\end{figure}

Graph Attention Networks (GAT) extend GCNs by incorporating an attention mechanism \cite{velickovic2017graph}. Instead of treating all neighbors equally, GATs compute attention scores between nodes by applying a shared linear transformation, followed by a dot product operation and a softmax normalization to determine the importance of each neighbor. The final node representation is obtained through a weighted sum of its neighbors' features. This dynamic weighting enhances expressiveness, enabling GATs to handle varying neighborhood sizes and adapt to dynamic graph structures. Unlike traditional GCNs, which rely on predefined adjacency matrices, GATs can learn important relationships directly from data, improving performance in node classification, graph classification, and link prediction. A sample illustration of weight and attention computation in GAT is shown in (\textbf{Fig. \ref{fig:gat-weight}}), adapted from \cite{velickovic2017graph}.

\subsubsection{GCN-GAT Hybrid Architecture}
The architecture combines Graph Convolution Networks (GCNs) and Graph Attention Networks (GATs) to utilize their strengths for processing graph-structured data. Initially, the GCN layers capture local neighborhood information by aggregating and propagating node-level features through graph convolution, modeling spatial relationships within the graph. Following this, GAT layers use an attention mechanism to assign different weights to neighboring nodes, allowing the model to focus on the most relevant features, thus enhancing its feature representation.

\begin{figure}[htbp]
  \centering
  \includegraphics[width=\textwidth,trim={0mm 7cm 0mm 3cm}, clip]{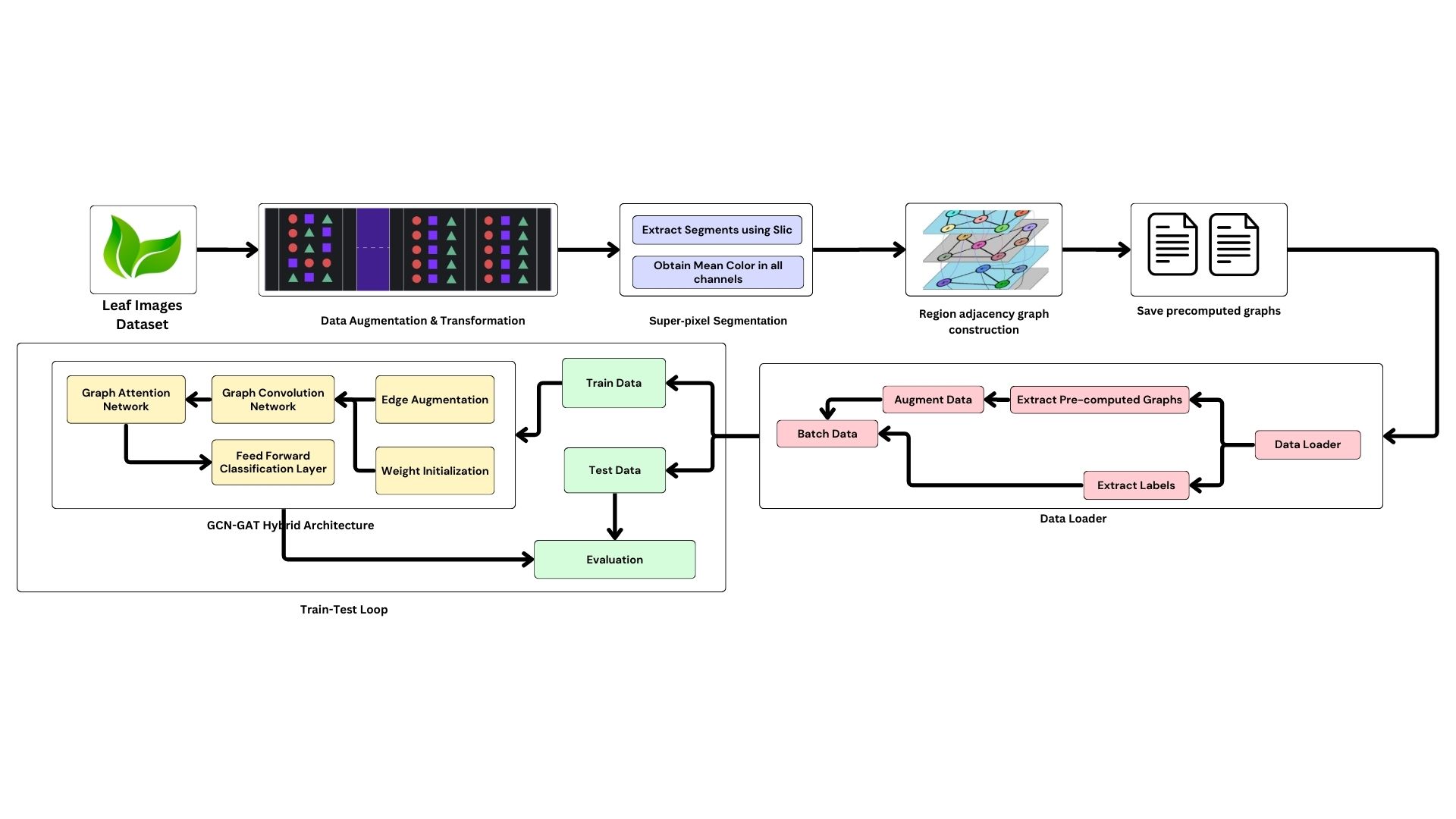}
  \caption{Architecture Diagram and Workflow of the Architecture}
  \label{fig:architecture-diagram}
\end{figure}

In addition, a classifier layer with LeakyReLU activation is then used to make predictions of the leaf conditions based on the learned representations. \textbf{Figure \ref{fig:architecture-diagram}} illustrates the workflow of the methodology.

To enhance model robustness and prevent overfitting, an edge augmentation strategy is employed. This stochastic approach introduces randomness by randomly adding or removing edges in the graph during training, encouraging the model to generalize better and reducing its reliance on specific graph structures. The algorithm for edge augmentation is represented in \textbf{Algorithm \ref{algo:edge-augmentation}}.

\begin{algorithm}[H]
  
  \KwData{Edge index: $\mathbf{E}$, Edge augmentation probability: $p$}
  \KwResult{Augmented edge index: $\mathbf{E}_{aug}$}
  Initialize: $\mathbf{E}_{aug} \gets \mathbf{E}$\;
  \If{random probability < $p$}{
    Add edge: $\mathbf{e}_{new} \gets$ random edge\;
    $\mathbf{E}_{aug} \gets \mathbf{E}_{aug} \cup \mathbf{e}_{new}$\;
  }
  \If{random probability < $p$}{
    Remove edge: $\mathbf{e}_{remove} \gets$ random edge from $\mathbf{E}_{aug}$\;
    $\mathbf{E}_{aug} \gets \mathbf{E}_{aug} \setminus \mathbf{e}_{remove}$\;
  }
  \Return{$\mathbf{E}_{aug}$}
  \caption{Edge Augmentation}
  \label{algo:edge-augmentation}
\end{algorithm}

To address the challenges of vanishing gradients and promote more efficient training, the He initialization method has been adopted \cite{he2015delving}. This method helps maintain consistent variance in the activations of each layer, enabling faster convergence and improving gradient flow throughout the network. The He initialization can be mathematically expressed as shown in \textbf{Equation \ref{eq:he-initialization}}.

\begin{equation}
    W \sim \mathcal{U}\left(-\sqrt{\frac{6}{n_{\text{in}}}}, \sqrt{\frac{6}{n_{\text{in}}}}\right) \label{eq:he-initialization}
\end{equation}
\begin{itemize}
    \item \(W\) represents the weight matrix.
    \item \(n_{\text{in}}\) is the number of input units (neurons) in the layer.
    \item The distribution is a uniform distribution, denoted by \(\mathcal{U}\), with bounds \(\pm \sqrt{\frac{6}{n_{\text{in}}}}\).
\end{itemize}
\subsection{Loss \& Evaluation Metrics}
Loss functions are mathematical functions used to measure how far predictions of th emodel deviate from the actual values. A higher loss indicates that the model is making more errors, while a lower loss suggests better predictive performance.

The architecture utilizes the cross-entropy loss function, which quantifies the difference between the predicted probability distribution and the true distribution.

\subsubsection{Cross-Entropy Loss}

The cross-entropy loss function is widely used in classification problems to quantify the deviation between the predicted probability distribution and the actual distribution. It helps to optimize the model by penalizing incorrect predictions more heavily. The mathematical representation of the cross-entropy loss function is provided in \textbf{Equation \ref{eq:cross-entropy loss}}.

\begin{equation}
     \text{{Cross-Entropy Loss}} = -\sum_{i=1}^{N} \sum_{j=1}^{C} y_{ij} \log(p_{ij}) \label{eq:cross-entropy loss}
    \end{equation}
    \begin{itemize}
      \item \(y\) represents the true distribution of the labels.
      \item \(\hat{y}\) represents the predicted distribution of the labels.
    \end{itemize}
    
Evaluation metrics are employed to assess the performance and quality of statistical or deep-learning models. To evaluate the hybrid model, the following metrics have been utilized:
\subsubsection{Accuracy}  
Accuracy represents the percentage of correctly classified leaf disease samples from the total number of leaf samples in the dataset. It measures the proportion of predictions that match the actual labels for each test image, reflecting the performance of the model in predicting the correct disease class label. Accuracy can be expressed as shown in \textbf{Equation \ref{eq:accuracy}}.

\begin{equation}
\label{eq:accuracy}
\text{Accuracy} = \frac{\text{Correct Predictions}}{\text{Total Predictions}} \times 100
\end{equation}

\subsubsection{Precision}  
Precision quantifies the reliability of the model when predicting a leaf as diseased. It specifically focuses on the cases where the model predicts the presence of a disease, evaluating how often these predictions are correct.  

The formula for calculating precision is given in \textbf{Equation \ref{eq:precision}}.

\begin{equation}
\label{eq:precision}
\text{Precision} = \frac{\text{True Positives}}{\text{True Positives} + \text{False Positives}}
\end{equation}

\subsubsection{Recall}  
Recall measures the ability of the model to correctly identify all diseased leaves from the total number of leaves that actually have the disease. It highlights the capabaility of the model to capture as many true positives as possible. The formula for calculating recall is presented in \textbf{Equation \ref{eq:recall}}.

\begin{equation}
\label{eq:recall}
\text{Recall} = \frac{\text{True Positives}}{\text{True Positives} + \text{False Negatives}}
\end{equation}

\subsubsection{F1-Score}  
The F1-score combines precision and recall into a single metric, providing a balanced evaluation of the performance of the model for each disease class. It accounts for both the ability of the model to detect diseased leaves and its accuracy in making predictions.  

The formula for calculating the F1-score is given in \textbf{Equation \ref{eq:f1-score}}.

\begin{equation} 
\label{eq:f1-score}
\text{F1-score} = \frac{2 \times \text{Precision} \times \text{Recall}}{\text{Precision} + \text{Recall}} 
\end{equation}

\subsubsection{Confusion Matrix}

The confusion matrix provides a detailed breakdown of the predictions of the model, categorizing them into various outcomes. For each class, it shows the number of samples that are correctly or incorrectly classified. This provides a comprehensive view of the performance of the model, highlighting both its accurate predictions and the areas where misclassifications occur.
    \[
    C = 
    \begin{bmatrix}
        TN & FP \\
        FN & TP \\
    \end{bmatrix}
    \]
Where:
\begin{itemize}
    \item \( TP \) signifies the count of true positives for class \(i\).
    \item \( TN \) signifies the count of true negatives for class \(i\).
    \item \( FP \) signifies the count of false positives for class \(i\).
    \item \( FN \) signifies the count of false negatives for class\(i\).
\end{itemize}

In multi-class classification, this matrix is extended such that each row and column corresponds to a class, providing a more granular evaluation of the performance of the model across all classes.

\section{Results and Discussions}
The datasets were evaluated using three distinct model architectures: Graph Convolution Networks (GCN), Graph Attention Networks (GAT), and the hybrid architecture, GCN-GAT. The performance of these models was analyzed to understand their effectiveness in leaf disease detection across diverse datasets. All models were trained for 100 epochs, and their performance was evaluated using standard evaluation metrics, including Precision, Recall, F1-Score, and Accuracy, alongside the analysis of loss and accuracy curves.The key training parameters for the experiments (determined through hyperparameter tuning) are summarized in \textbf{Table \ref{table:training-parameters}}.

\begin{table}[http]
  \centering
  \caption{Parameter settings}
  \begin{tabular}{| p{5cm}| p{5cm} | }
    \toprule
    \textbf{Parameter} & \textbf{Settings} \\
    \midrule
    Image Size & (128,128,3) \\
    Batch Size & 32 \\
    Learning Rate & 0.001 \\
    Optimizer & Adam \\
    Attention heads & 2 \\
    Hidden Layers & 512 \\
    GCN Layers & 2 \\
    GAT Layers & 2 \\
    \bottomrule
  \end{tabular}
  \label{table:training-parameters}
\end{table}

For the Apple Leaf dataset, the GCN model achieved an accuracy of 92.71\%, with a precision of 0.9265, recall of 0.9271, and an F1-Score of 0.9267. In comparison, the GAT model attained an accuracy of 82.03\%, with precision, recall, and F1-Score values of 0.8558, 0.8203, and 0.8202, respectively. The hybrid architecture (GCN-GAT) demonstrated exceptional performance, achieving an accuracy of 99.73\%, along with a precision of 0.9974, recall of 0.9974, and an F1-Score of 0.9974. The performance results of the different models on the Apple Leaf dataset are summarized in \textbf{Table \ref{table:apple-leaf-performance}}.
\begin{table}[htbp]
    \centering
    \renewcommand{\arraystretch}{1.3}
    \caption{\textbf{Performance Metrics for Apple Leaf Dataset}}
    \begin{tabular}{|p{3cm}|p{1.5cm}|p{1.8cm}|p{1.5cm}|p{1.5cm}|p{2cm}|}
        \hline
        \textbf{Model} & \textbf{Accuracy} & \textbf{Precision} & \textbf{Recall} & \textbf{F1-Score} & \textbf{Average Loss} \\
        \hline
        GCN & 0.9271 & 0.9265 & 0.9271 & 0.9267 & 0.1987 \\
        GAT & 0.8203 & 0.8558 & 0.8203 & 0.8202 & 0.8102 \\
        \textbf{GCN+GAT} & \textbf{0.9973} & \textbf{0.9974} & \textbf{0.9974} & \textbf{0.9974} & \textbf{0.0143} \\
        \hline
    \end{tabular}
    \label{table:apple-leaf-performance}
\end{table}

Building on these results, the models were subsequently evaluated on the Potato Leaf dataset to further assess their effectiveness. On this dataset, the GCN model achieved an accuracy of 94.77\%, with a precision of 0.9506, recall of 0.9478, and an F1-Score of 0.9475. Interestingly, the GAT model slightly underperformed compared to GCN, attaining an accuracy of 92.33\%, with a precision of 0.9298, recall of 0.9233, and an F1-Score of 0.9243. However, the hybrid GCN-GAT model once again stood out, delivering the highest performance with an accuracy of 98.11\%, a precision of 0.9811, recall of 0.9811, and an F1-Score of 0.9811. The performance metrics for the models on the Potato Leaf dataset are presented in \textbf{Table \ref{table:potato-leaf-performance}}, highlighting the comparison between GCN, GAT, and the hybrid GCN-GAT model.
\begin{table}[htbp]
    \centering
    \renewcommand{\arraystretch}{1.3}
    \caption{\textbf{Performance Metrics for Potato Leaf Dataset}}
    \begin{tabular}{|p{3cm}|p{1.5cm}|p{1.8cm}|p{1.5cm}|p{1.5cm}|p{2cm}|}
        \hline
        \textbf{Model} & \textbf{Accuracy} & \textbf{Precision} & \textbf{Recall} & \textbf{F1-Score} & \textbf{Average Loss} \\
        \hline
        GCN & 0.9477 & 0.9506 & 0.9478 & 0.9475 & 0.1447 \\
        GAT & 0.9233 & 0.9298 & 0.9233 & 0.9243 & 0.3902 \\
        \textbf{GCN+GAT} & \textbf{0.9811} & \textbf{0.9811} & \textbf{0.9811} & \textbf{0.9811} & \textbf{0.0548} \\
        \hline
    \end{tabular}
    \label{table:potato-leaf-performance}
\end{table}

Finally, the models were tested on the Sugarcane Leaf dataset, which posed unique challenges due to complex patterns present in the images. On this dataset, the GCN model achieved an accuracy of 64.63\%, with a precision of 0.6745, recall of 0.6464, and an F1-Score of 0.6399. The GAT model experienced a notable drop in performance, yielding an accuracy of 48.83\%, with precision, recall, and F1-Score values of 0.5360, 0.4884, and 0.4703, respectively. In contrast, the GCN-GAT hybrid model demonstrated its robustness, achieving a significantly higher accuracy of 91.03\%, with a precision of 0.9124, recall of 0.9104, and an F1-Score of 0.9101. These results are summarized in \textbf{Table \ref{table:sugarcane-leaf-performance}}.
\begin{table}[htbp]
    \centering
    \renewcommand{\arraystretch}{1.3}
    \caption{\textbf{Performance Metrics for Sugarcane Leaf Dataset}}
    \begin{tabular}{|p{3cm}|p{1.5cm}|p{1.8cm}|p{1.5cm}|p{1.5cm}|p{2cm}|}
        \hline
        \textbf{Model} & \textbf{Accuracy} & \textbf{Precision} & \textbf{Recall} & \textbf{F1-Score} & \textbf{Average Loss} \\
        \hline
        GCN & 0.6463 & 0.6745 & 0.6464 & 0.6399 & 0.9301 \\
        GAT & 0.4883 & 0.5360 & 0.4884 & 0.4703 & 1.2542 \\
        \textbf{GCN+GAT} & \textbf{0.9103} & \textbf{0.9124} & \textbf{0.9104} & \textbf{0.9101} & \textbf{0.2651} \\
        \hline
    \end{tabular}
    \label{table:sugarcane-leaf-performance}
\end{table}

\begin{figure}[!htbp]
    \centering
    \begin{subfigure}[b]{0.3\textwidth}
        \centering
        \includegraphics[width=\linewidth]{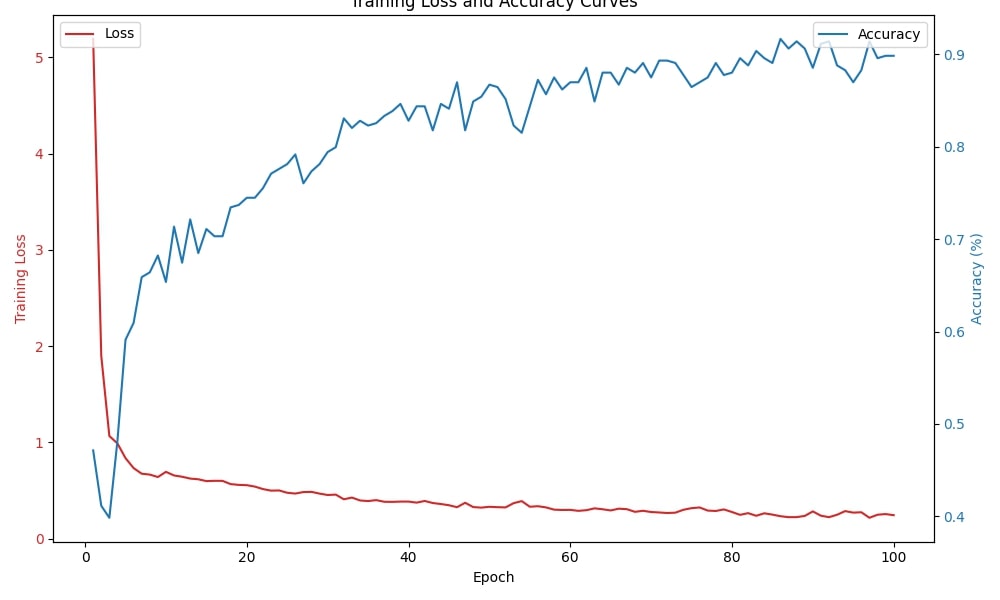}
        \caption{GCN - Apple Leaf Dataset}
        \label{fig:apple-gcn-training-curves}
    \end{subfigure}
    \hfill
    \begin{subfigure}[b]{0.3\textwidth}
        \centering
        \includegraphics[width=\linewidth]{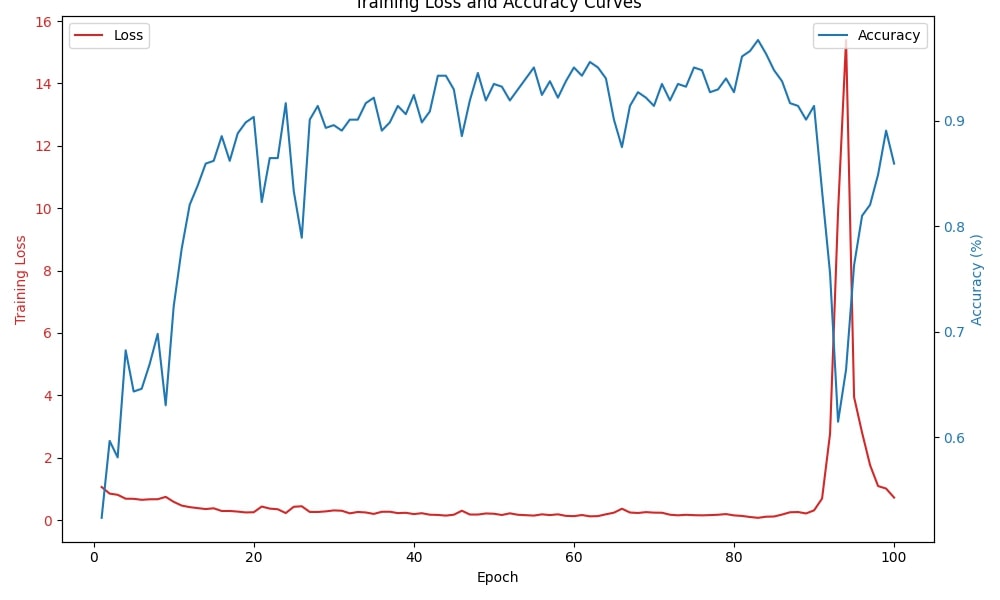}
        \caption{GAT - Apple Leaf Dataset}
        \label{fig:apple-gat-training-curves}
    \end{subfigure}
    \hfill
    \begin{subfigure}[b]{0.3\textwidth}
        \centering
        \includegraphics[width=\linewidth]{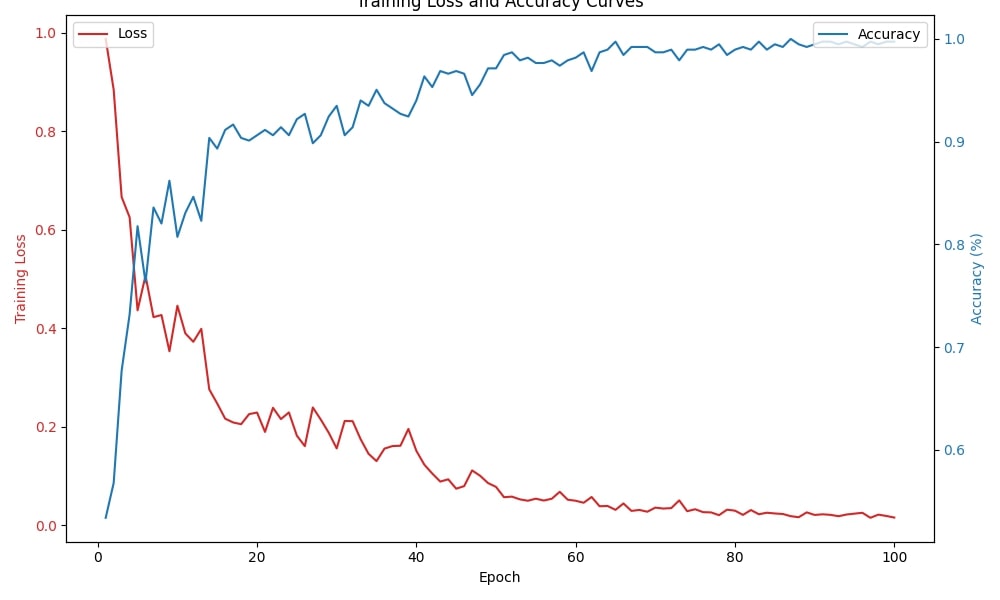}
        \caption{GCN-GAT Hybrid - Apple Leaf Dataset}
        \label{fig:apple-gcn-gat-training-curves}
    \end{subfigure}
    \caption{Training loss and accuracy curves for GCN, GAT, and GCN-GAT models on the Apple Leaf Disease dataset.}
    \label{fig:apple-leaf-disease-training-curves}
\end{figure}

To evaluate training performance across different datasets, the loss and accuracy curves were examined. For the Apple Leaf Disease dataset, the GCN model recorded an average loss of 0.1987, whereas the GAT model showed a higher loss of 0.8102. The hybrid GCN-GAT architecture outperformed both, achieving a minimal loss of 0.0143, showcasing its superior learning capability. The corresponding loss and accuracy curves are depicted in \textbf{Fig.\ref{fig:apple-leaf-disease-training-curves}}.

\begin{figure}[!htbp]
    \centering
    \begin{subfigure}[b]{0.3\textwidth}
        \centering
        \includegraphics[width=\linewidth]{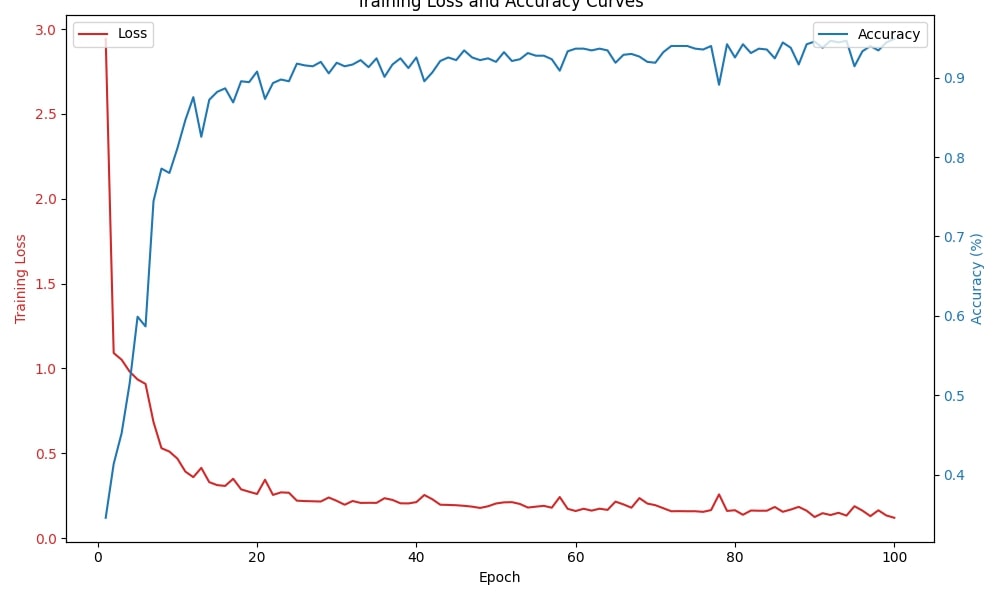}
        \caption{GCN - Potato Leaf Dataset}
        \label{fig:potato-gcn-training-curves}
    \end{subfigure}
    \hfill
    \begin{subfigure}[b]{0.3\textwidth}
        \centering
        \includegraphics[width=\linewidth]{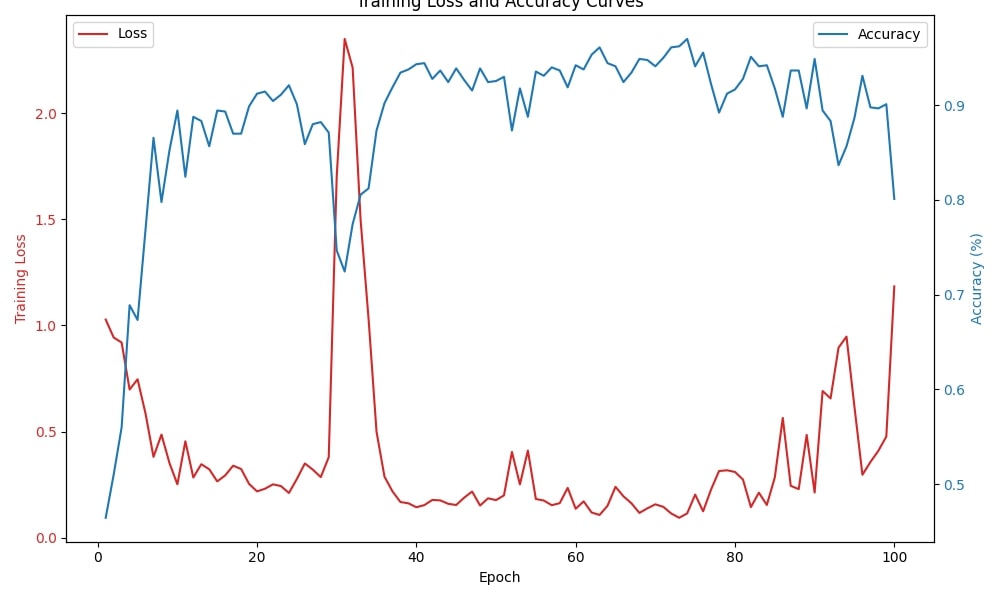}
        \caption{GAT - Potato Leaf Dataset}
        \label{fig:potato-gat-training-curves}
    \end{subfigure}
    \hfill
    \begin{subfigure}[b]{0.3\textwidth}
        \centering
        \includegraphics[width=\linewidth]{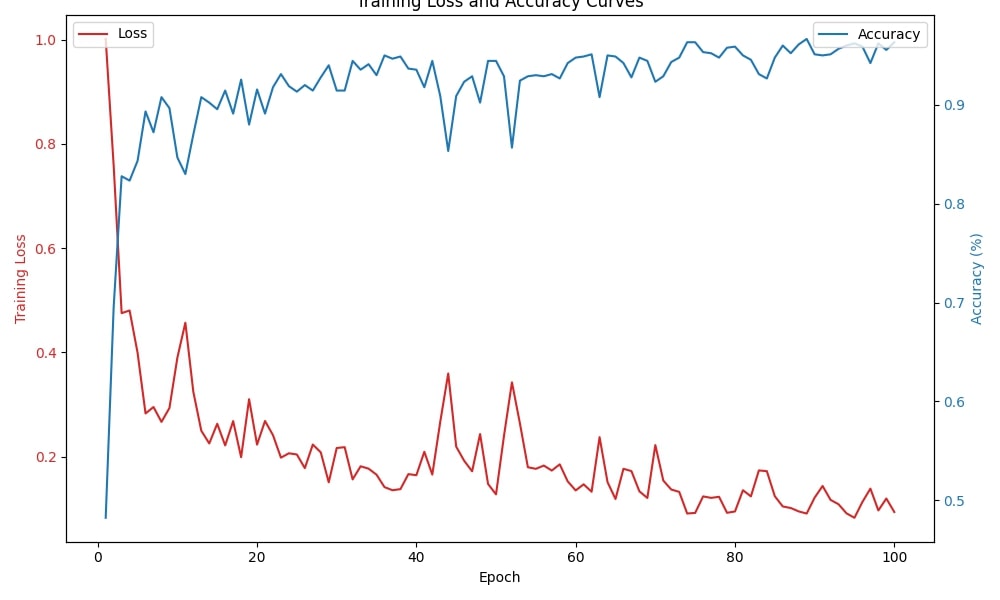}
        \caption{GCN-GAT Hybrid - Potato Leaf Dataset}
        \label{fig:potato-gcn-gat-training-curves}
    \end{subfigure}
    \caption{Training loss and accuracy curves for GCN, GAT, and GCN-GAT models on the Potato Leaf Disease dataset.}
    \label{fig:potato-leaf-disease-training-curves}
\end{figure}

On the Potato Leaf Disease dataset, the GCN model achieved a loss of 0.1447, while the GAT model demonstrated slightly higher efficiency with a loss of 0.3902. The hybrid architecture maintained its advantage, attaining the lowest loss of 0.0548. The respective training curves are illustrated in \textbf{Fig. \ref{fig:potato-leaf-disease-training-curves}}.

\begin{figure}[!htbp]
    \centering
    \begin{subfigure}[b]{0.3\textwidth}
        \centering
        \includegraphics[width=\linewidth]{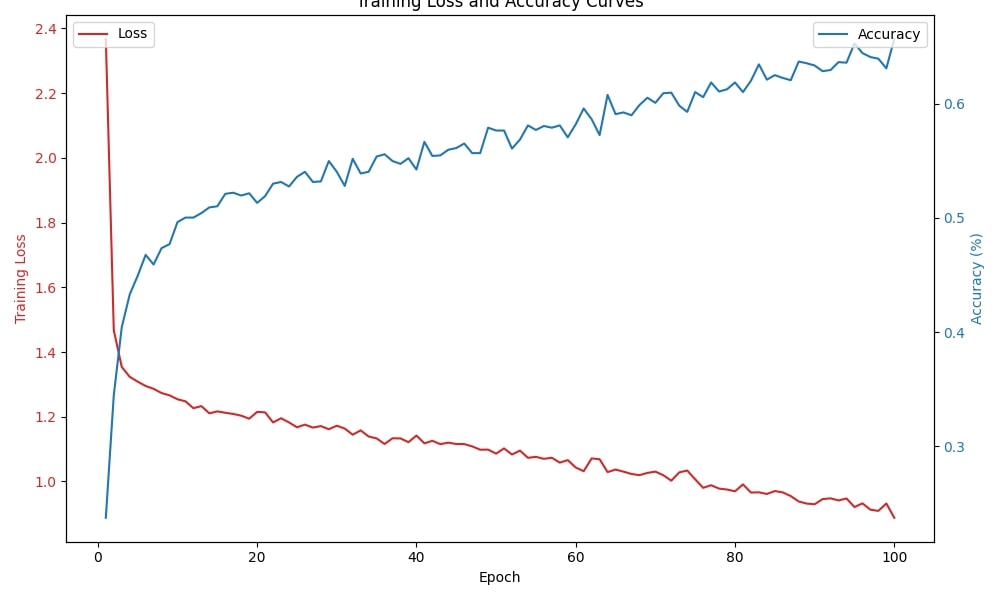}
        \caption{GCN - Sugarcane Leaf Dataset}
        \label{fig:sugarcane-gcn-training-curves}
    \end{subfigure}
    \hfill
    \begin{subfigure}[b]{0.3\textwidth}
        \centering
        \includegraphics[width=\linewidth]{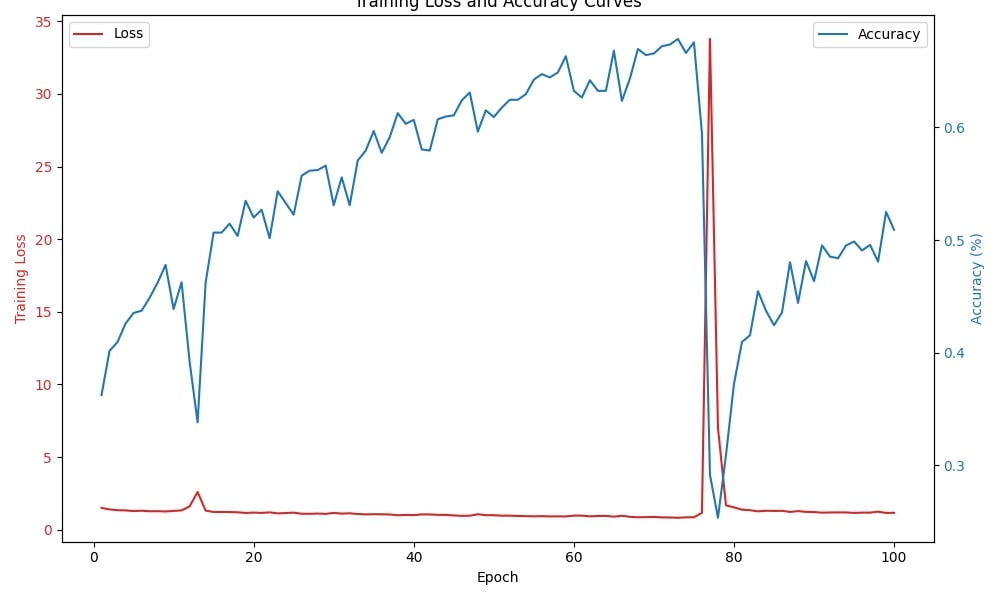}
        \caption{GAT - Sugarcane Leaf Dataset}
        \label{fig:sugarcane-gat-training-curves}
    \end{subfigure}
    \hfill
    \begin{subfigure}[b]{0.3\textwidth}
        \centering
        \includegraphics[width=\linewidth]{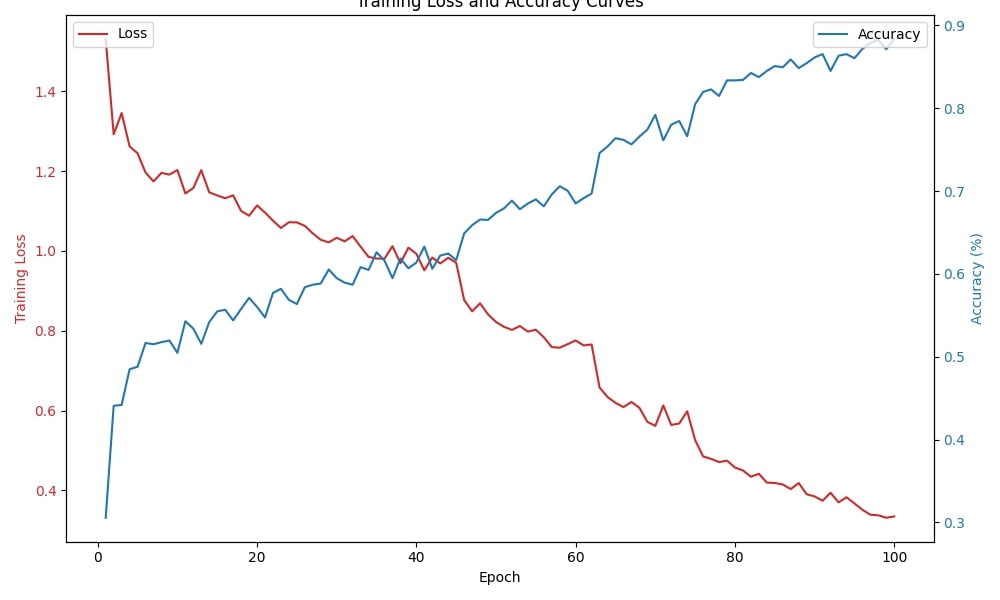}
        \caption{GCN-GAT Hybrid - Sugarcane Leaf Dataset}
        \label{fig:sugarcane-gcn-gat-training-curves}
    \end{subfigure}
    \caption{Training loss and accuracy curves for GCN, GAT, and GCN-GAT models on the Sugarcane Leaf Disease dataset.}
    \label{fig:sugarcane-leaf-disease-training-curves}
\end{figure}

Similarly, for the Sugarcane Leaf Disease dataset, the GCN model experienced a loss of 0.9301, with the GAT model facing greater difficulty, recording a loss of 1.2542. The hybrid architecture once again proved to be the most effective, with a significantly reduced loss of 0.2651. The training performance curves for this dataset are presented in \textbf{Fig. \ref{fig:sugarcane-leaf-disease-training-curves}}.

To assess the classification performance per class in each dataset, the respective confusion matrices of individual architectures were inspected and analyzed.

\begin{figure}[!htbp]
    \centering
    \begin{subfigure}[b]{0.3\textwidth}
        \centering
        \includegraphics[width=\linewidth]{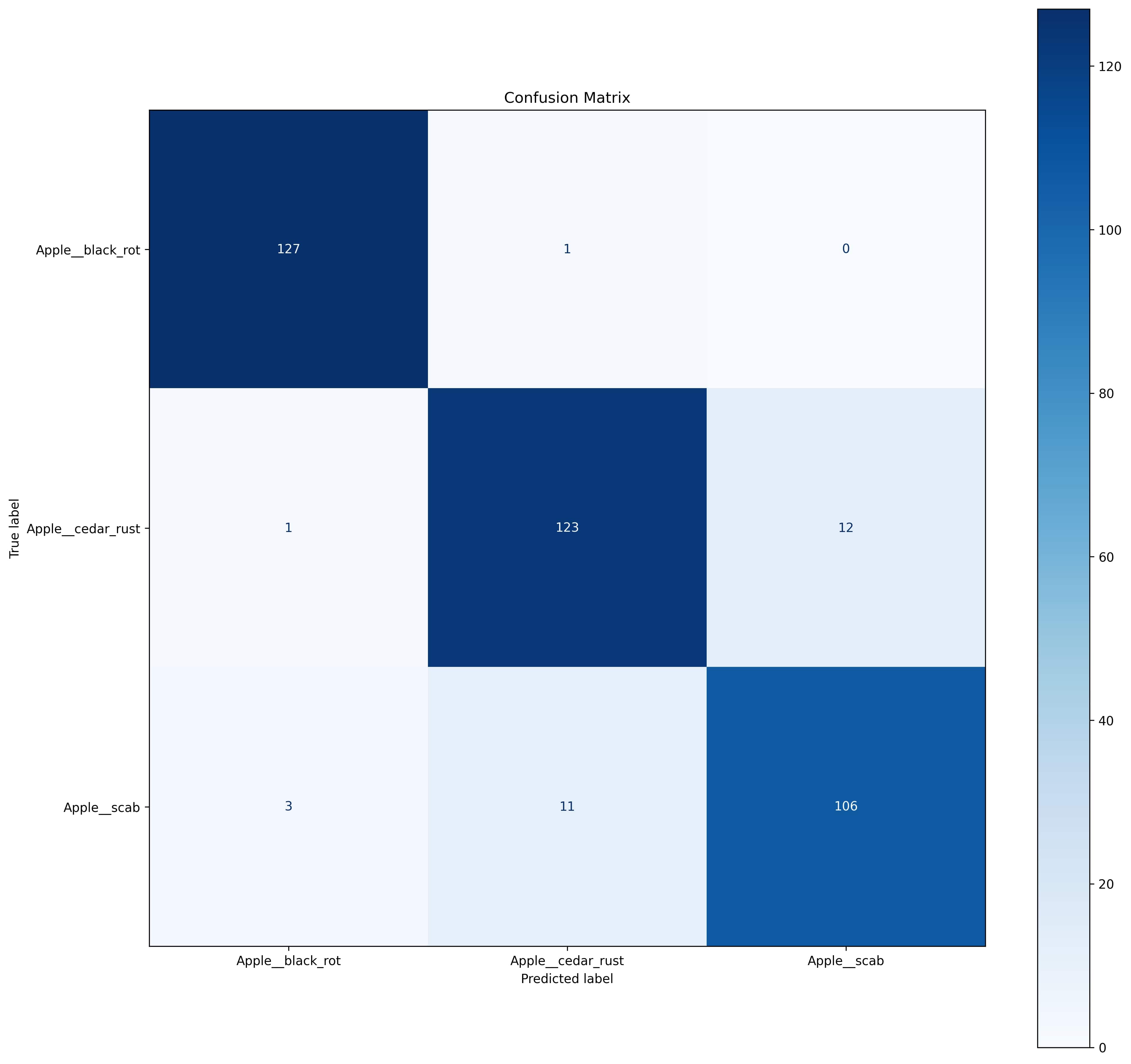}
        \caption{GCN - Apple Leaf Dataset}
        \label{fig:apple-gcn-confusion-matrix}
    \end{subfigure}
    \hfill
    \begin{subfigure}[b]{0.3\textwidth}
        \centering
        \includegraphics[width=\linewidth]{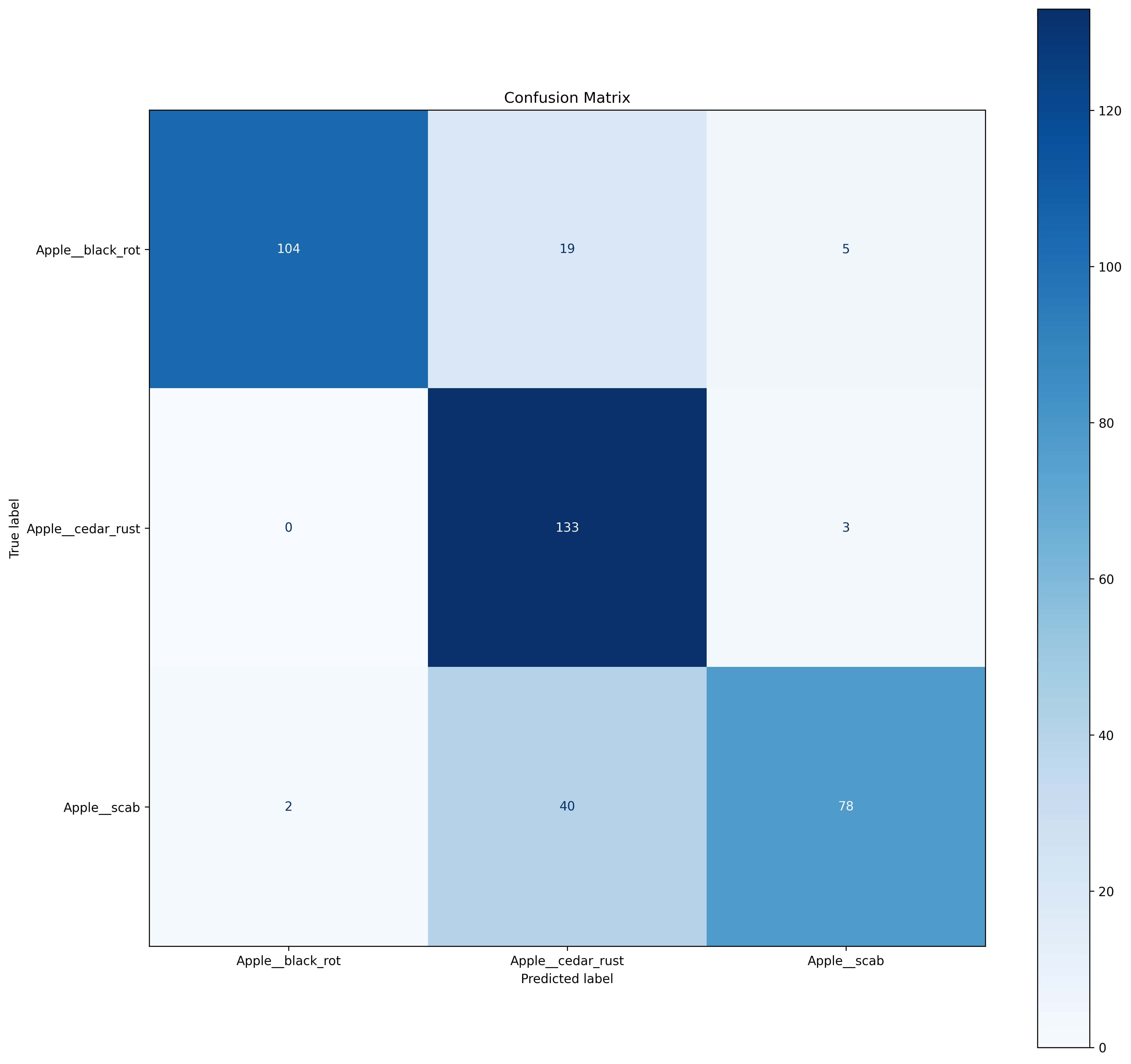}
        \caption{GAT - Apple Leaf Dataset}
        \label{fig:apple-gat-confusion-matrix}
    \end{subfigure}
    \hfill
    \begin{subfigure}[b]{0.3\textwidth}
        \centering
        \includegraphics[width=\linewidth]{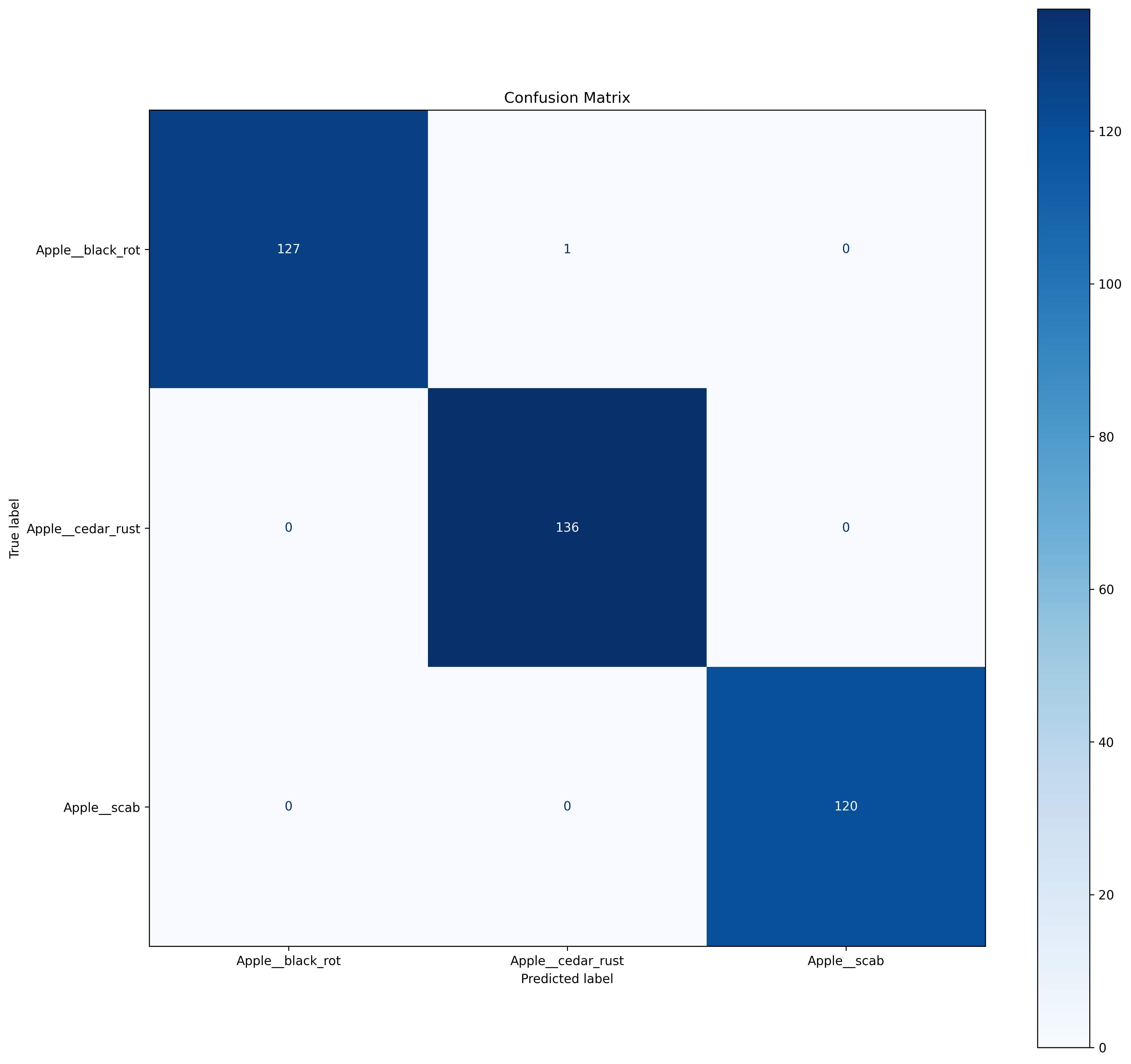}
        \caption{GCN-GAT Hybrid - Apple Leaf Dataset}
        \label{fig:apple-gcn-gat-confusion-matrix}
    \end{subfigure}
    \caption{Confusion matrices for GCN, GAT, and GCN-GAT models on the Apple Leaf Disease dataset.}
    \label{fig:apple-leaf-disease-confusion-matrix}
\end{figure}

In the apple leaf disease dataset, the GCN model effectively classified black rot apple leaves but misclassified 13 out of 136 samples in rust and 14 out of 120 in scab. The GAT model, however, underperformed across all classes, resulting in a higher frequency of misclassifications. In contrast, the GCN-GAT Hybrid model demonstrated superior performance, reducing misclassifications to 1 out of 128 in Black Rot leaf condition. The corresponding confusion matrices are shown in \textbf{Fig.\ref{fig:apple-leaf-disease-confusion-matrix}}.

\begin{figure}[!htbp]
    \centering
    \begin{subfigure}[b]{0.3\textwidth}
        \centering
        \includegraphics[width=\linewidth]{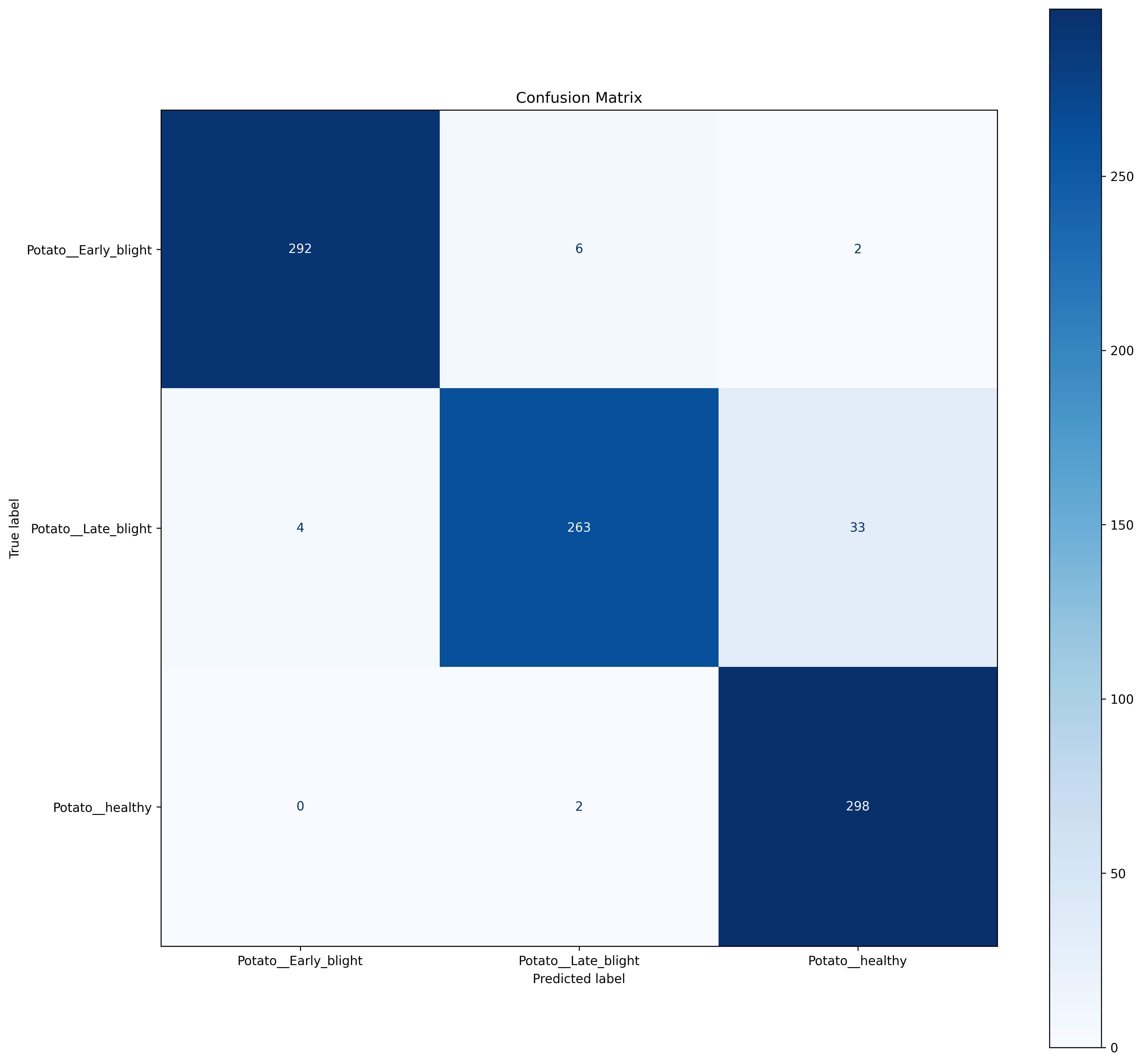}
        \caption{GCN - Potato Leaf Dataset}
        \label{fig:potato-gcn-confusion-matrix}
    \end{subfigure}
    \hfill
    \begin{subfigure}[b]{0.3\textwidth}
        \centering
        \includegraphics[width=\linewidth]{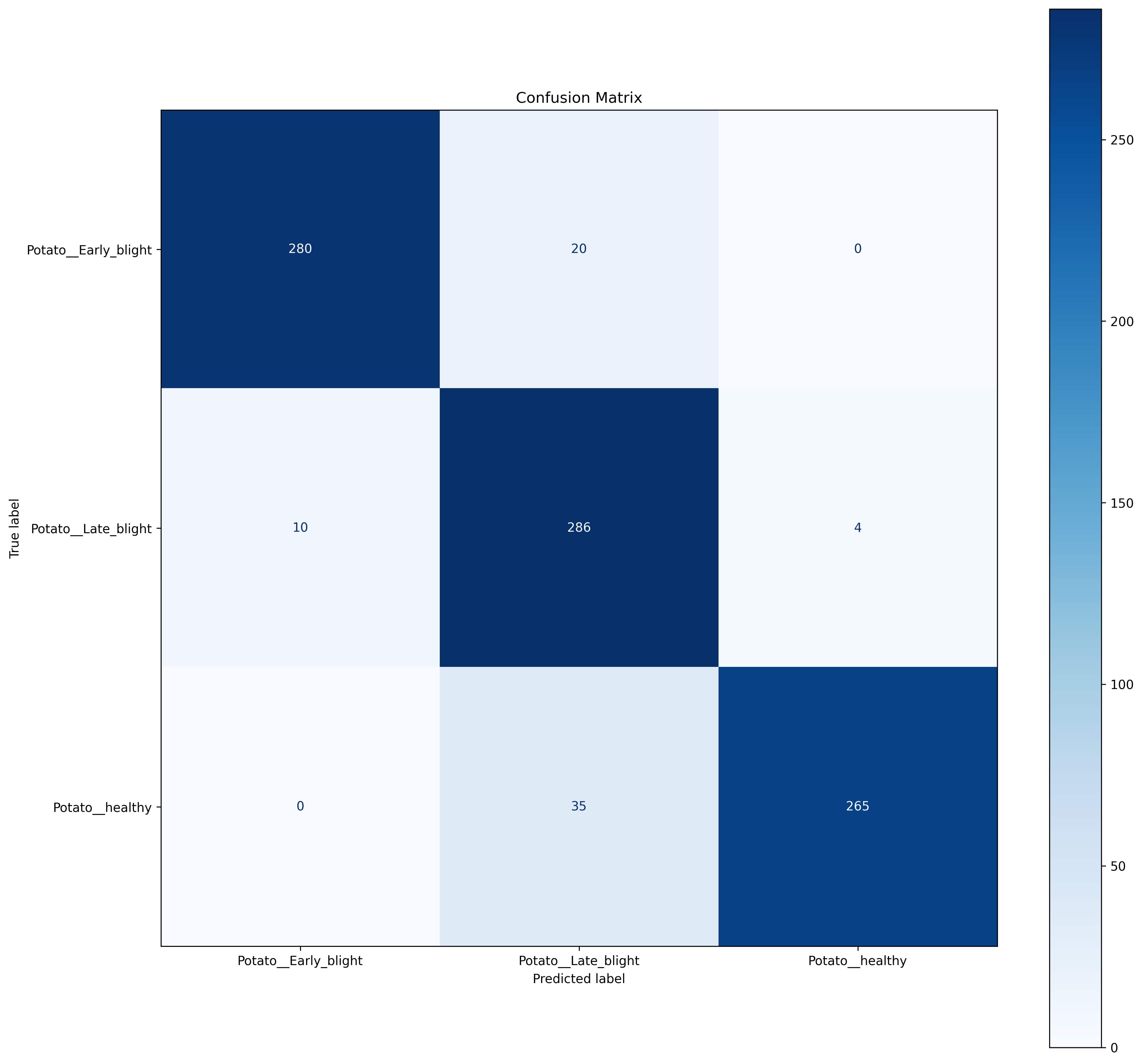}
        \caption{GAT - Potato Leaf Dataset}
        \label{fig:potato-gat-confusion-matrix}
    \end{subfigure}
    \hfill
    \begin{subfigure}[b]{0.3\textwidth}
        \centering
        \includegraphics[width=\linewidth]{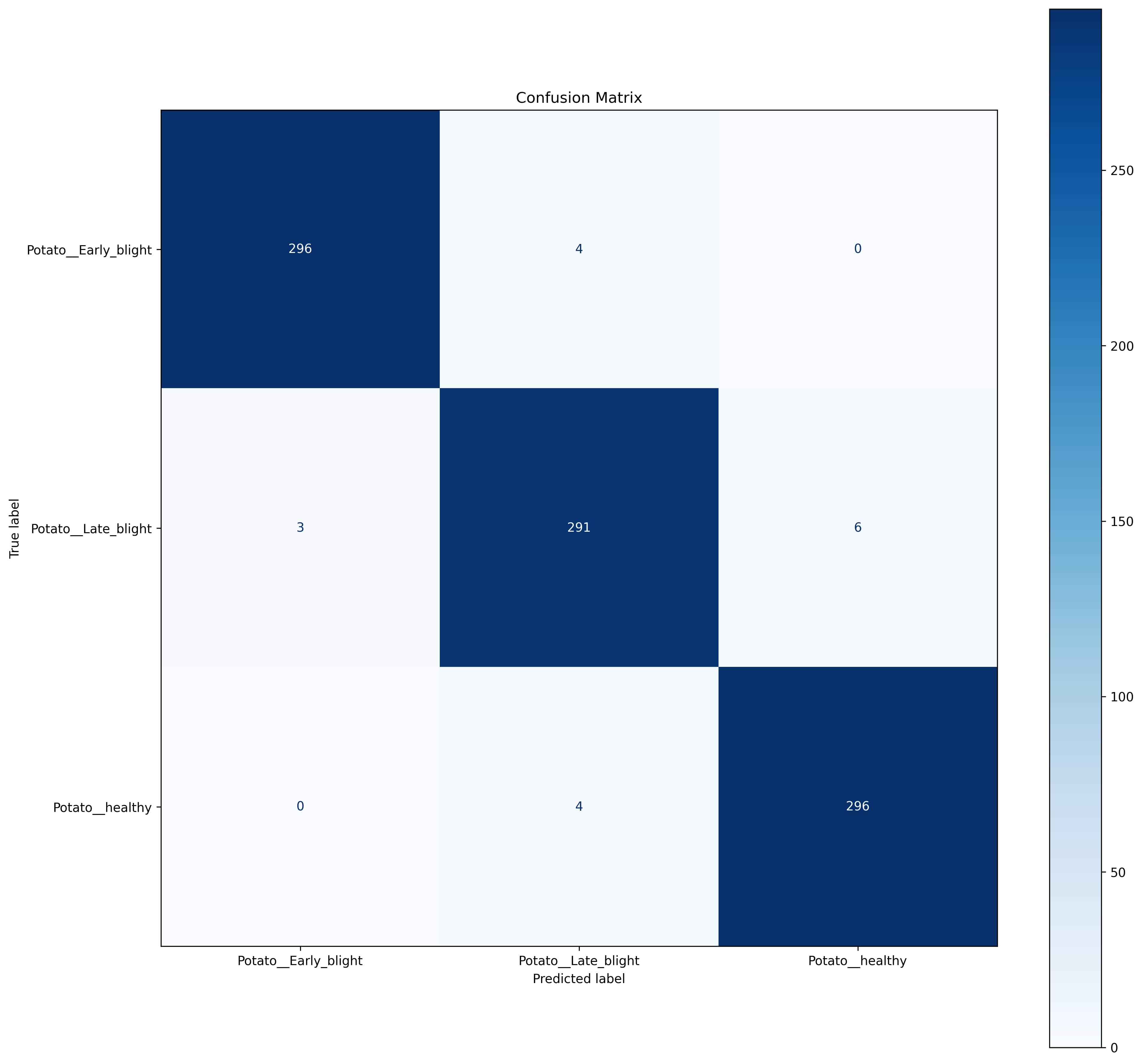}
        \caption{GCN-GAT Hybrid - Potato Leaf Dataset}
        \label{fig:potato-gcn-gat-confusion-matrix}
    \end{subfigure}
    \caption{Confusion matrices for GCN, GAT, and GCN-GAT models on the Potato Leaf Disease dataset.}
    \label{fig:potato-leaf-disease-confusion-matrix}
\end{figure}

On the potato leaf disease dataset, the GCN model misclassified 37 out of 300 samples in Late Blight conditions. Although the GAT model showed consistent performance across all classes, the GCN-GAT Hybrid model outperformed both, correctly classifying 883 out of 900 samples. The confusion matrices for each model are presented in \textbf{Fig.\ref{fig:potato-leaf-disease-confusion-matrix}}.

\begin{figure}[!htbp]
    \centering
    \begin{subfigure}[b]{0.3\textwidth}
        \centering
        \includegraphics[width=\textwidth]{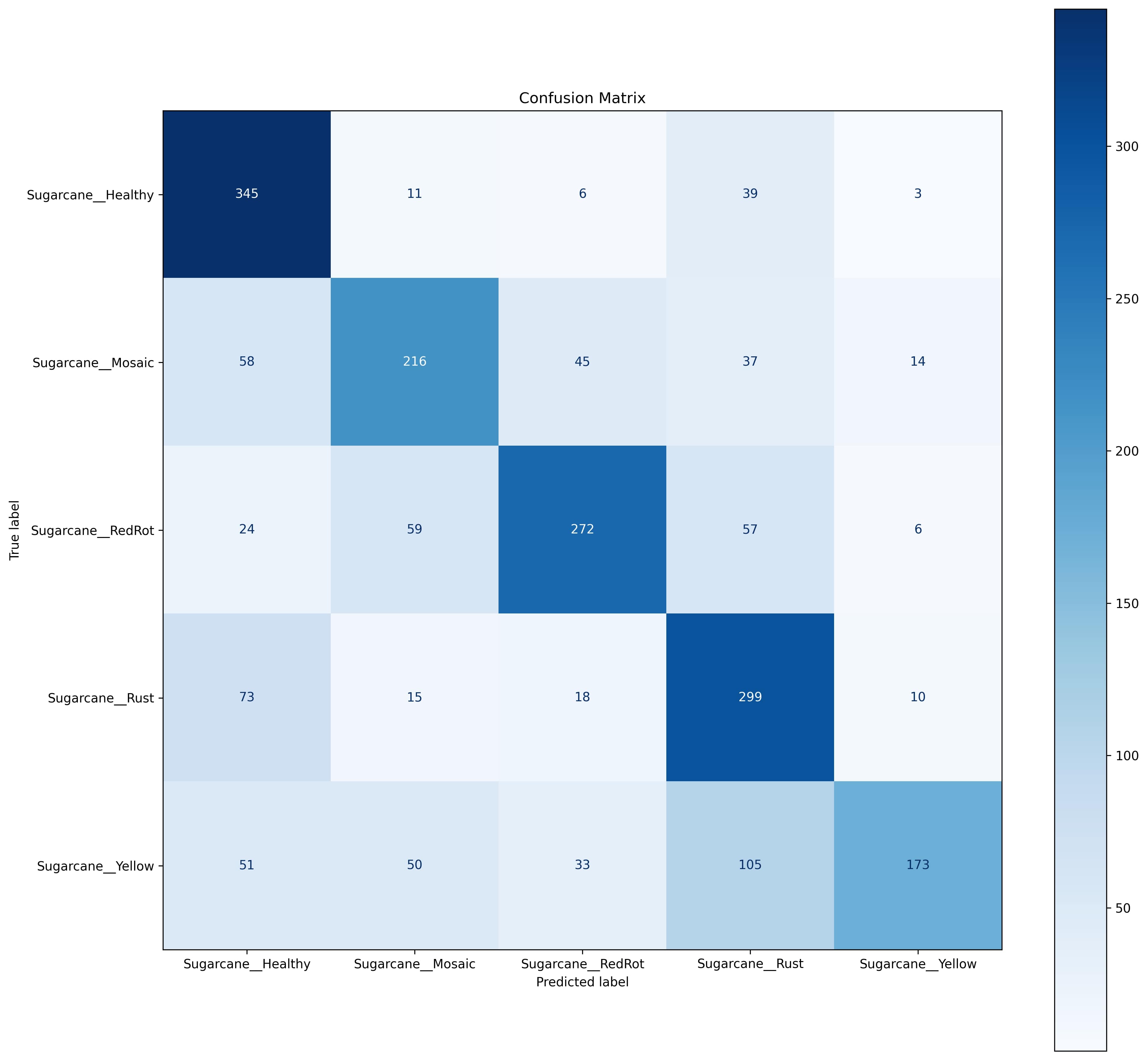}
        \caption{GCN - Sugarcane Leaf Dataset}
        \label{fig:sugarcane-gcn-confusion-matrix}
    \end{subfigure}
    \hfill
    \begin{subfigure}[b]{0.3\textwidth}
        \centering
        \includegraphics[width=\textwidth]{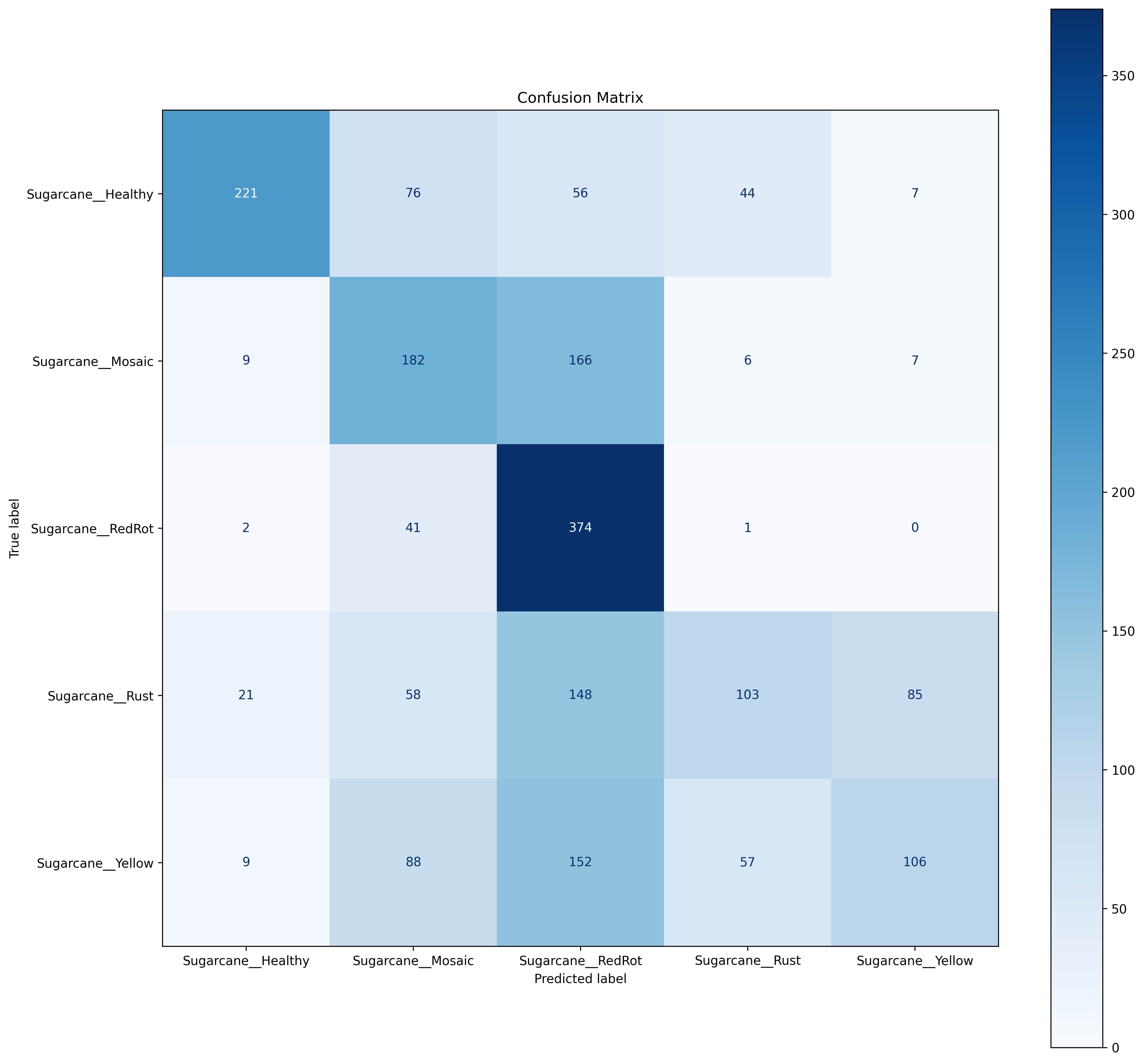}
        \caption{GAT - Sugarcane Leaf Dataset}
        \label{fig:sugarcane-gat-confusion-matrix}
    \end{subfigure}
    \hfill
    \begin{subfigure}[b]{0.3\textwidth}
        \centering
        \includegraphics[width=\textwidth]{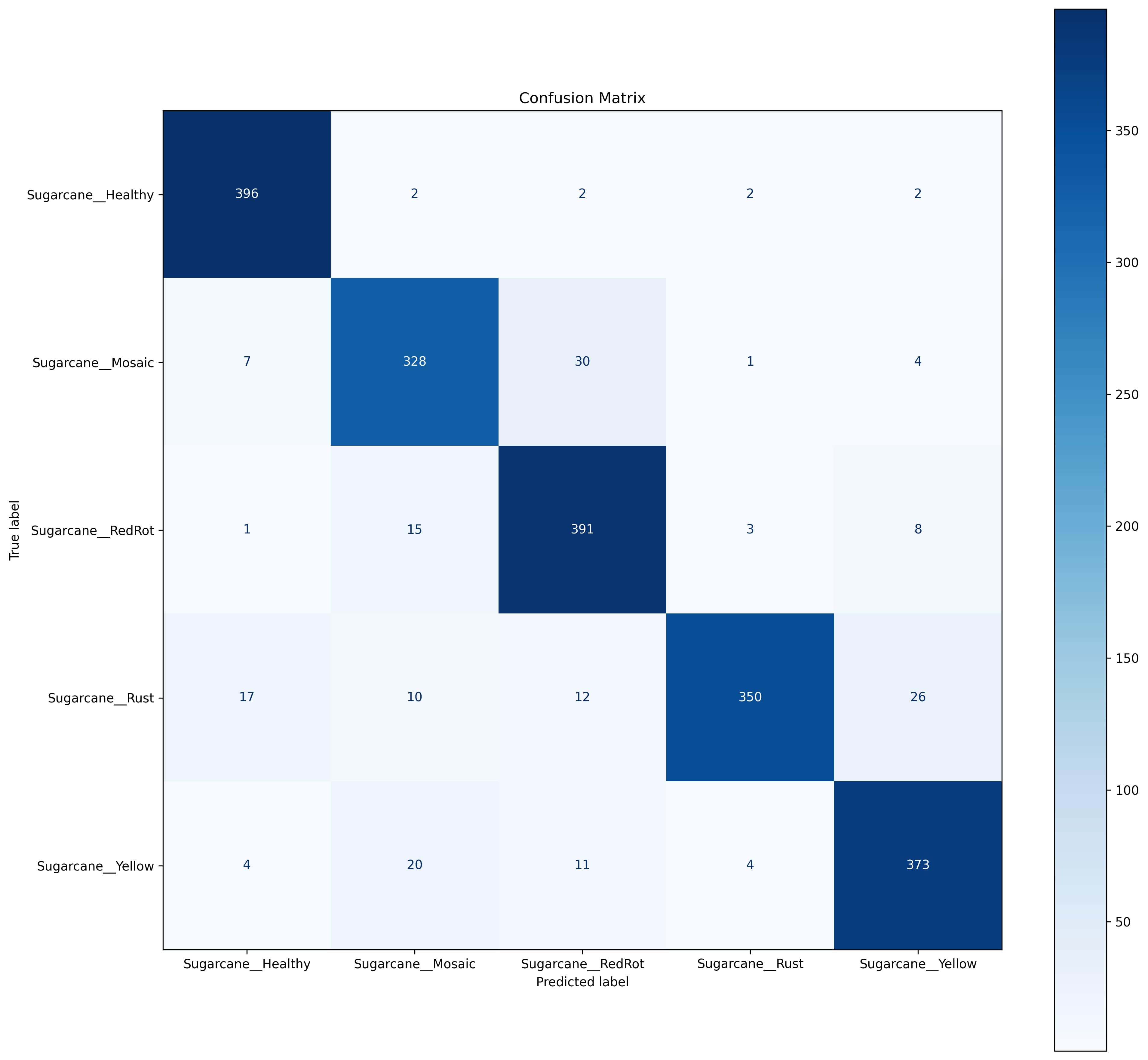}
        \caption{GCN-GAT Hybrid - Sugarcane Leaf Dataset}
        \label{fig:sugarcane-gcn-gat-confusion-matrix}
    \end{subfigure}
    \caption{Confusion matrices for GCN, GAT, and GCN-GAT models on the Sugarcane Leaf Disease dataset.}
    \label{fig:sugarcane-leaf-disease-confusion-matrix}
\end{figure}

Finally, on the sugarcane leaf disease dataset, the GCN model exhibited varying performance across different classes, performing well in some but misclassifying more samples in others. The GAT model showed poor performance overall, making random predictions across classes. In contrast, the GCN-GAT Hybrid model, despite some misclassifications in Mosaic and Mosaic leaf conditions, outperformed both individual models. \textbf{Fig.\ref{fig:sugarcane-leaf-disease-confusion-matrix}} presents the confusion matrix for the sugarcane dataset.

\begin{figure}[!htbp]
    \centering
    \includegraphics[width=0.9\textwidth]{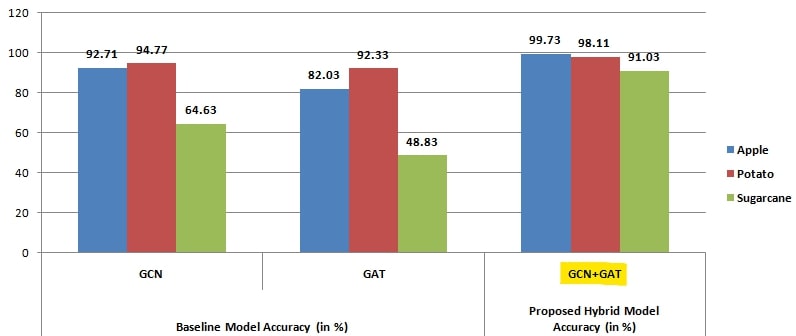}
    \caption{Graph depicting the accuracy of GCN, GAT, and the proposed GCN+GAT models for the classification of diseases on Apple, Sugarcane, and Potato leaves datasets.}
    \label{fig:overall-performance}
\end{figure}

\textbf{Fig. \ref{fig:overall-performance}} highlights the overall accuracy of all the three architectures across multiple datasets, demonstrating the superior performance of the GCN-GAT Hybrid model in consistently achieving higher classification accuracy compared to GCN and GAT.

\section{Conclusion}
An architecture using the GCN-GAT hybrid model was analyzed for leaf disease detection, leveraging the spatial feature-capturing capabilities of Graph Convolution Networks (GCN) and the feature prioritization strengths of Graph Attention Networks (GAT). The model was trained and evaluated on three diverse datasets: apple leaves (3 classes), potato leaves (3 classes), and sugarcane leaves (5 classes), representing varied leaf conditions. To highlight the superiority of the hybrid model, its performance was thoroughly analyzed and compared against the standalone GCN and GAT models.

The hybrid model exhibited superior performance across the datasets. On the apple dataset, it achieved a precision of 0.9974, recall of 0.9974, and an F1-score of 0.9974, with an average loss of 0.0143. For the potato dataset, the model achieved precision, recall, and F1-scores of 0.9811, 0.9811, and 0.9811, respectively, with an average loss of 0.0548. Despite the complexity of the sugarcane dataset, the model maintained robust results, achieving precision, recall, and F1-scores of 0.9124, 0.9104, 0.9101 each, with an average loss of 0.2651. These metrics underscore the adaptability and effectiveness of the hybrid model, with average accuracies of 99.74\% for apple leaves, 98.11\% for potato leaves, and 91.03\% for sugarcane leaves.

For comparison, the standalone GCN and GAT models delivered lower metrics across all datasets. On the apple dataset, the GCN model achieved a precision of 0.9443, recall of 0.9438, and an F1-score of 0.9436, with an average loss of 0.0345, while the GAT model achieved a precision of 0.9521, recall of 0.9519, and an F1-score of 0.9517, with an average loss of 0.0302. On the potato dataset, GCN recorded precision, recall, and F1-scores of 0.9134, 0.9129, and 0.9127, with an average loss of 0.0386, whereas GAT achieved precision, recall, and F1-scores of 0.9291, 0.9286, and 0.9284, with an average loss of 0.0331. For the sugarcane dataset, GCN and GAT showed relatively lower performance, with GCN achieving a precision of 0.8112, recall of 0.8108, and an F1-score of 0.8105 (average loss: 0.0564), and GAT achieving precision, recall, and F1-scores of 0.8293, 0.8290, and 0.8288 (average loss: 0.0517).

In summary, the GCN-GAT hybrid model significantly outperformed the standalone GCN and GAT models across all datasets, demonstrating its ability to effectively balance spatial feature extraction and feature prioritization. The inclusion of both models' strengths enabled the hybrid architecture to achieve higher accuracy and lower loss, making it a robust solution for leaf disease classification.

Beyond current implementations, future work could focus on optimizing the computational efficiency of the hybrid model to facilitate its deployment on low-power devices. The integration of the proposed model into a real-time agricultural framework has the potential to revolutionize disease detection systems, ensuring healthier crops, reducing losses, and contributing to crop security.

\section*{Conflict of Interest Statement}

To the best of our knowledge, there are no conflicts of interest.

\bibliographystyle{unsrt}
\bibliography{references}

\end{document}